\documentclass{article}
\usepackage{arxiv}

\usepackage{graphicx}
\usepackage{amsmath}
\usepackage{amssymb}
\usepackage{booktabs}

\usepackage{url}
\usepackage{multirow,makecell, rotating}
\usepackage{booktabs}
\usepackage{tikz}
\usepackage{caption}
\usepackage{graphbox}
\usepackage{comment}
\usepackage{epsfig}
\usepackage{bm}
\usepackage{soul}
\usepackage{color}
\usepackage{subcaption}

\usepackage[pagebackref=true,breaklinks=true,letterpaper=true,colorlinks,bookmarks=false]{hyperref} 

\usepackage[capitalize]{cleveref}
\crefname{section}{Sec.}{Secs.}
\Crefname{section}{Section}{Sections}
\Crefname{table}{Table}{Tables}
\crefname{table}{Tab.}{Tabs.}

\begin{document}

\title{Back to the Feature: Classical 3D Features are (Almost) All You Need for 3D Anomaly Detection}

\author{Eliahu Horwitz, Yedid Hoshen\\
  School of Computer Science and Engineering\\
  The Hebrew University of Jerusalem, Israel\\
  \url{http://www.vision.huji.ac.il/3d_ads/}\\
  \texttt{\{eliahu.horwitz, yedid.hoshen\}@mail.huji.ac.il} \\
}

\maketitle

\begin{figure*}[h]
    \centering
   \begin{tabular}{@{\hskip1pt}c@{\hskip1pt}c@{\hskip1pt}c@{\hskip1pt}c}
            \includegraphics[width=0.22\linewidth]{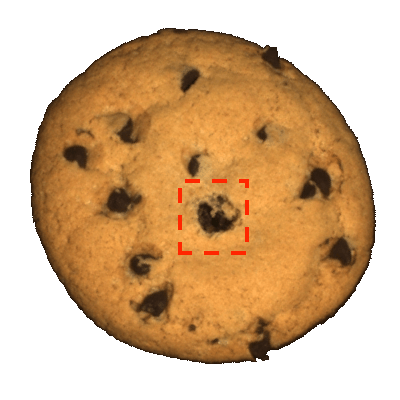} & 
            \includegraphics[width=0.22\linewidth]{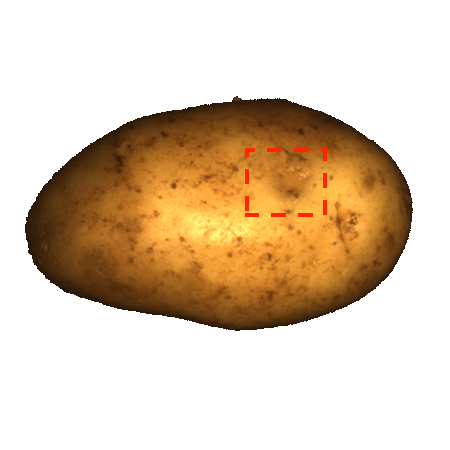} &
            \includegraphics[width=0.22\linewidth]{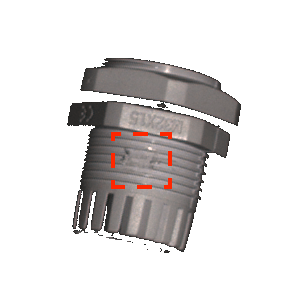} &
            \includegraphics[width=0.22\linewidth]{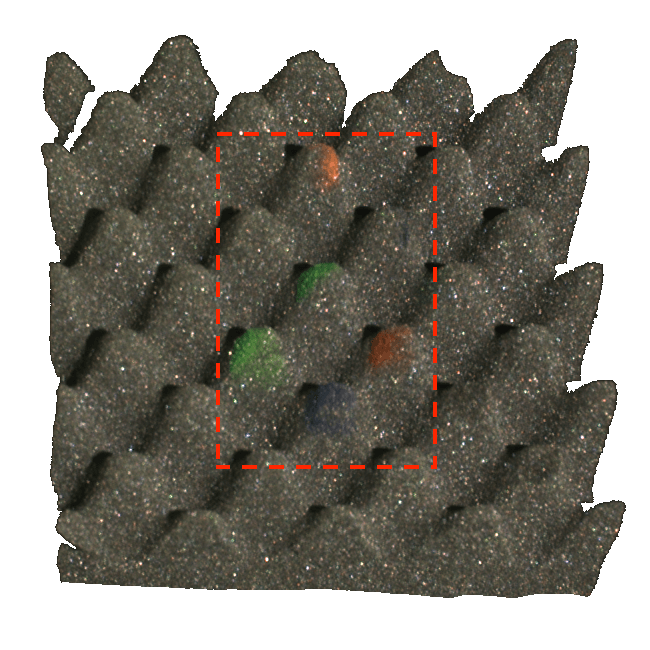}  \\
            \includegraphics[width=0.22\linewidth]{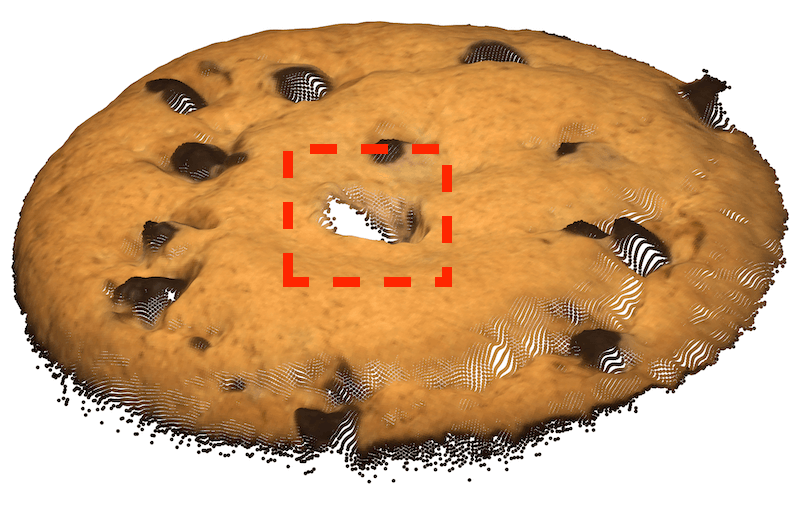} & 
            \includegraphics[width=0.22\linewidth]{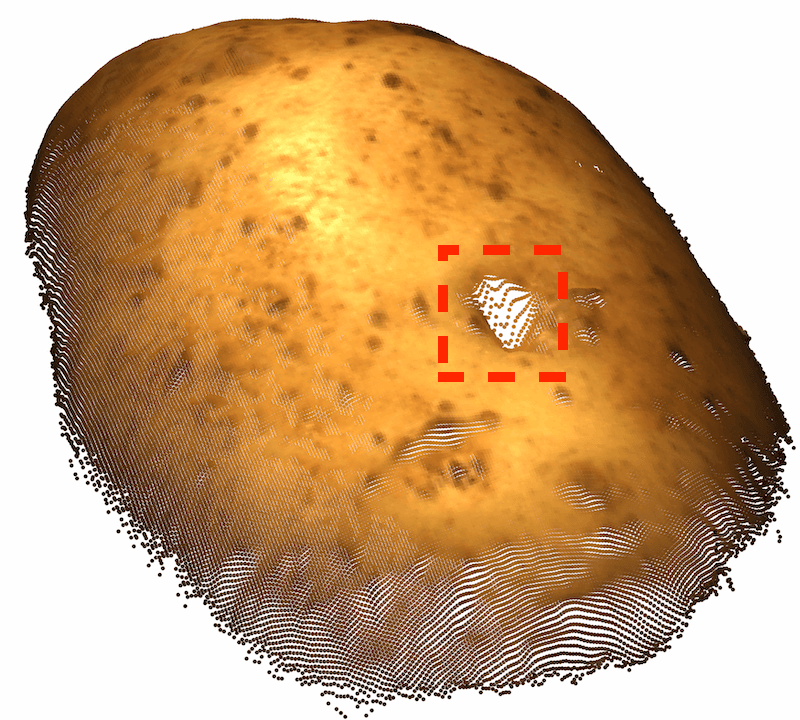} &
            \includegraphics[width=0.22\linewidth]{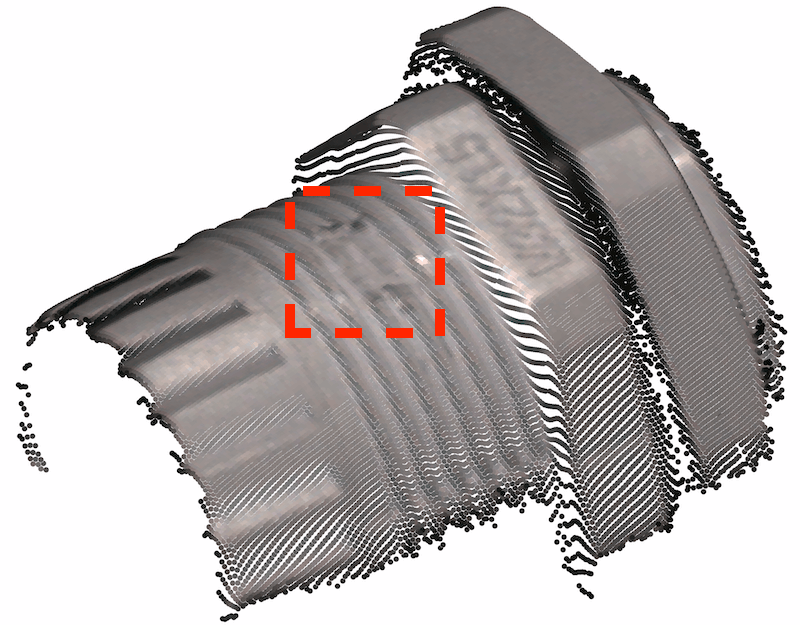} & 
            \includegraphics[width=0.22\linewidth]{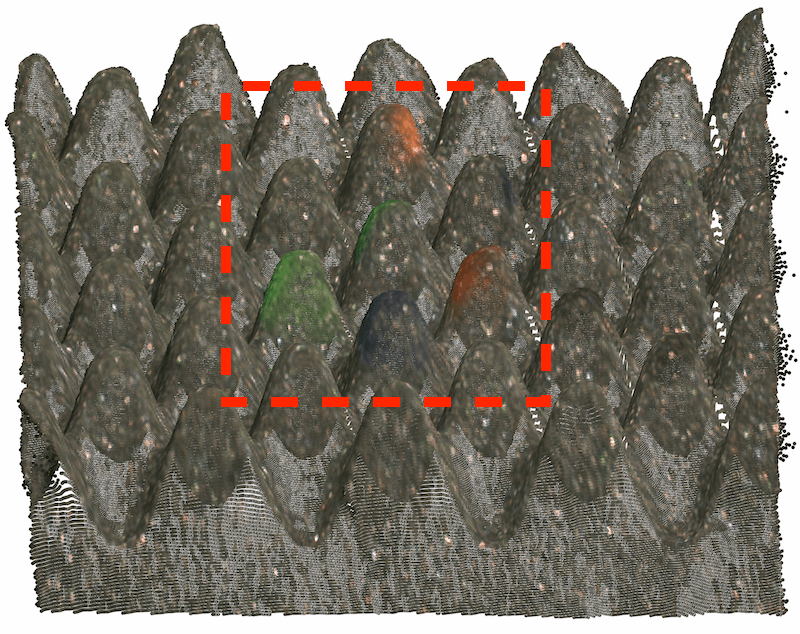} \\
        \end{tabular}
\caption{\textit{\textbf{Color and 3D - better together:}} Objects can be anomalous in both shape and texture. Some shape anomalies are easily detected as sharp deformations in the 3D shape (marked in red, two leftmost objects - cookie and potato). In such cases, color is ineffective; the anomalies cannot be detected in top-row views. Conversely, 3D information often cannot identify texture anomalies e.g., deformation in cable gland, color of foam (two rightmost objects). These anomalies are easily detected in the 2D color images}
\label{fig:rgb_v_d}
\end{figure*}

\begin{abstract}
Despite significant advances in image anomaly detection and segmentation, few methods use 3D information. We utilize a recently introduced 3D anomaly detection dataset to evaluate whether or not using 3D information is a lost opportunity. First, we present a surprising finding: standard color-only methods outperform all current methods that are explicitly designed to exploit 3D information.
This is counter-intuitive as even a simple inspection of the dataset shows that color-only methods are insufficient for images containing geometric anomalies. This motivates the question: how can anomaly detection methods effectively use 3D information? We investigate a range of shape representations including hand-crafted and deep-learning-based; we demonstrate that rotation invariance plays the leading role in the performance.
We uncover a simple 3D-only method that beats all recent approaches while not using deep learning, external pre-training datasets, or color information.
As the 3D-only method cannot detect color and texture anomalies, we combine it with color-based features, significantly outperforming previous state-of-the-art. Our method, dubbed BTF (Back to the Feature) achieves pixel-wise ROCAUC: $99.3\%$ and PRO: $96.4\%$ on MVTec 3D-AD.

\end{abstract}

\section{Introduction}

Although 3D understanding is fundamental to computer vision, it has typically not been considered by image anomaly detection and segmentation approaches, probably because of the lack of suitable datasets.
To encourage research into 3D anomaly detection and segmentation (AD\&S), the MVTec 3D-AD \cite{mvtec3d} dataset was recently introduced alongside several baseline methods for 3D AD\&S.
However, despite the existence of a 3D AD\&S dataset, the role of 3D information, as opposed to color-only, is still unclear.
We conduct a careful study seeking answers to several questions:
\begin{enumerate}
    \item Do current 3D AD\&S methods truly outperform state-of-the-art 2D methods on 3D data?
    \item Is 3D information \textit{potentially} useful for AD\&S?
    \item What are the key properties of successful 3D AD\&S representations?
    \item Are there complementary benefits from using 3D shape and color modalities?
\end{enumerate}

As very few previous image AD\&S methods have used 3D information, we conducted a preliminary investigation of baseline methods on the MVTec-3D dataset. Perhaps surprisingly, color-only methods (e.g. PatchCore \cite{patchcore}) outperform all current 3D AD\&S methods by a wide margin. 
Next, we ask whether 3D information is potentially 
useful for AD\&S. Encouragingly, we find that several types of anomalies go undetected when using color-only information (see Fig.~\ref{fig:rgb_v_d}~leftmost~two,~top~row).
In the bottom row, we present another view of the same objects, rendered using the 3D point cloud, where the anomalies are easily detected\footnote[1]{Note that the black back-plane of the images was removed for visualization purposes. In some cases, this removal is only possible given the 3D information (e.g. We cannot tell apart the hole from the chocolate by looking only at the color image, 3D information is needed).}

Having shown that 3D information is often needed for AD\&S, our goal is to identify effective 3D representations for AD\&S.  We investigate a broad range of hand-crafted and deep representations and find that rotation invariance is key for 3D AD\&S. \textit{Our surprising result: a classical, handcrafted 3D point cloud descriptor outperforms all other current methods, including learning-based representations.}

Notwithstanding the previous results, it is clear that color information is helpful. E.g., we present some examples from MVTec 3D-AD where the anomaly is much clearer in the color than in the shape (Fig.~\ref{fig:rgb_v_d}, rightmost two examples). 
This motivates our final approach, \textit{BTF} (Back to the Feature) which combines 3D and color to achieve the best-recorded result on the MVTec 3D-AD dataset by a very wide margin ($99.3\%$ \textit{Pixel-wise ROCAUC}, $96.4\%$ \textit{PRO}, and $87.3\%$ \textit{Image ROCAUC}). 

Our main contributions in this paper are:
\begin{itemize}
    \item[$\bullet$] Conducting a thorough analysis of the important and unexplored field of anomaly detection and segmentation for images with color and 3D information.
    \item[$\bullet$] Identifying that current 2D representations significantly outperform 3D representations on 3D data.
    \item[$\bullet$] Discovering that rotation invariant representations are key for 3D AD\&S.
    \item[$\bullet$] Proposing BTF, a method that combines handcrafted 3D representations (FPFH) with a deep, color-based method (PatchCore), outperforming the state-of-the-art by a wide margin. 
\end{itemize}


\begin{figure*}[t]
\begin{tabular}{@{\hskip1pt}c@{\hskip1pt}c@{\hskip1pt}c@{\hskip1pt}c@{\hskip1pt}c}

\rotatebox[origin=c]{90}{Input} &
\raisebox{-0.5\height}{\includegraphics[width=0.25\linewidth]{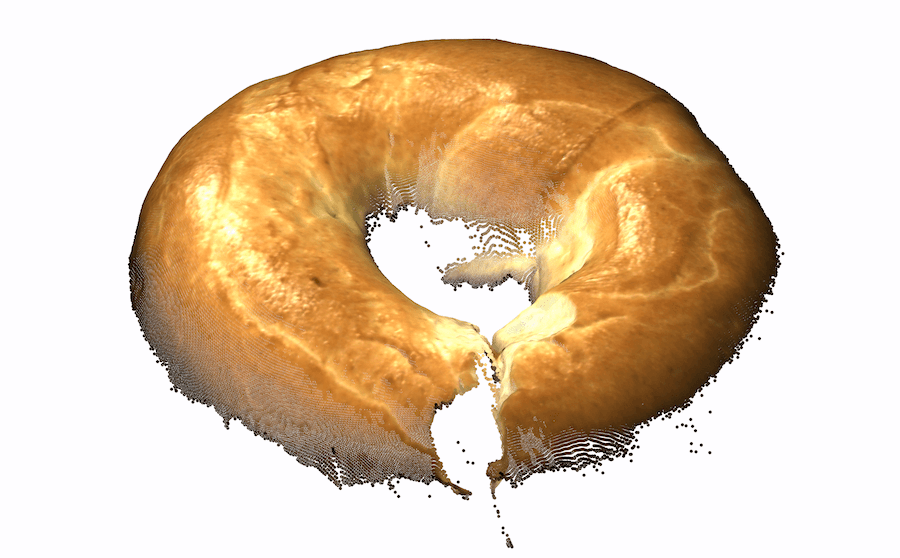}} & 
\raisebox{-0.5\height}{\includegraphics[width=0.25\linewidth]{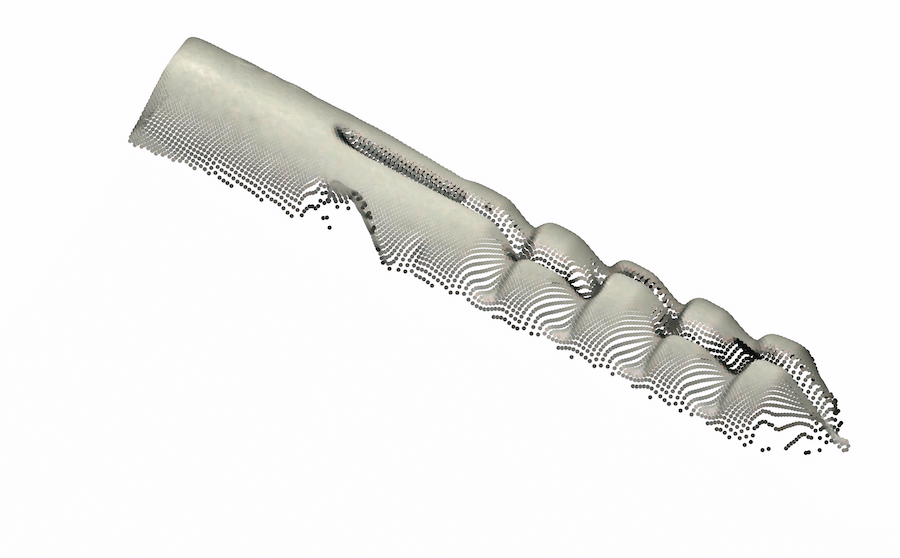}} & 
\raisebox{-0.5\height}{\includegraphics[width=0.25\linewidth]{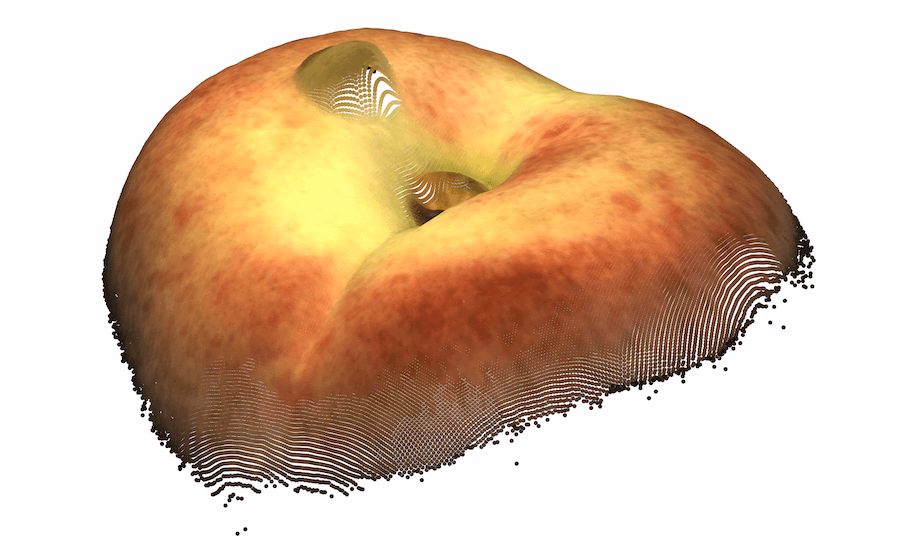}} & 
\raisebox{-0.5\height}{\includegraphics[width=0.25\linewidth]{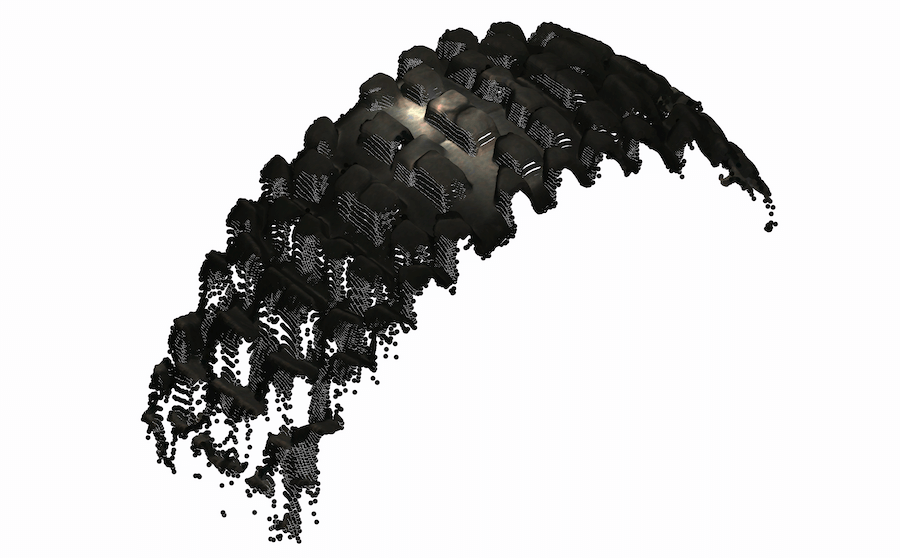}}\\
\rotatebox[origin=c]{90}{GT} &
\raisebox{-0.5\height}{\includegraphics[width=0.25\linewidth]{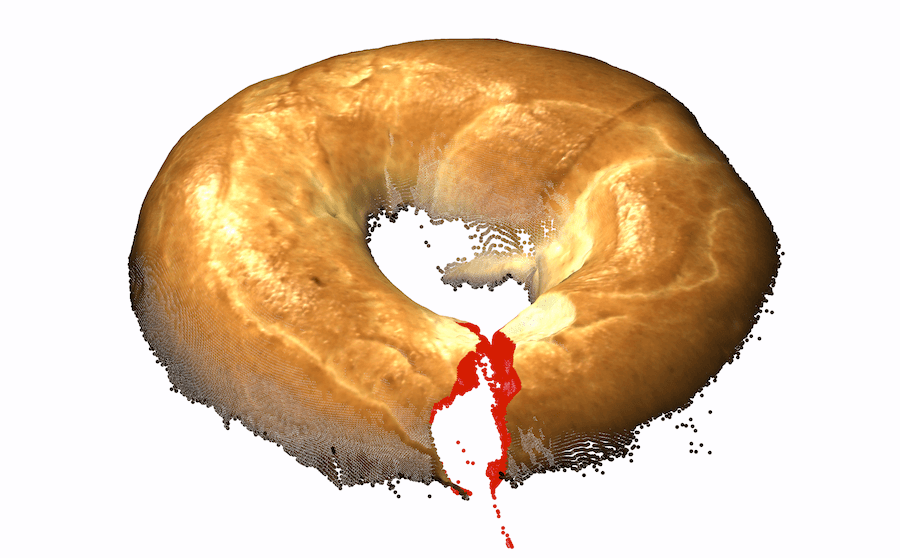}} & 
\raisebox{-0.5\height}{\includegraphics[width=0.25\linewidth]{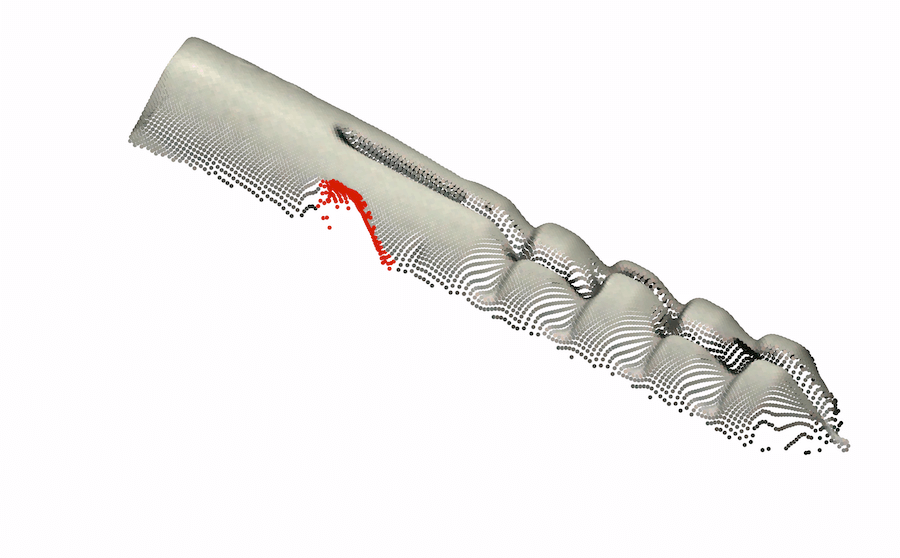}} &
\raisebox{-0.5\height}{\includegraphics[width=0.25\linewidth]{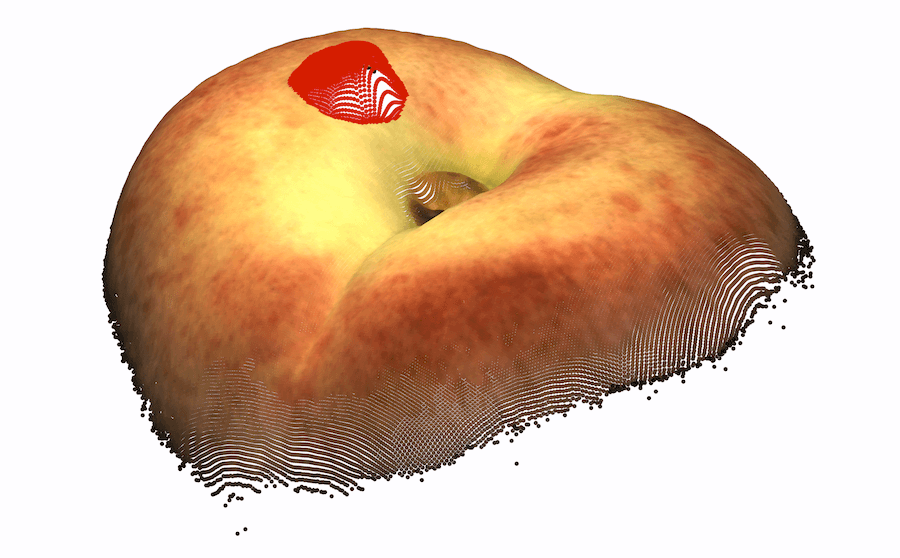}} & 
\raisebox{-0.5\height}{\includegraphics[width=0.25\linewidth]{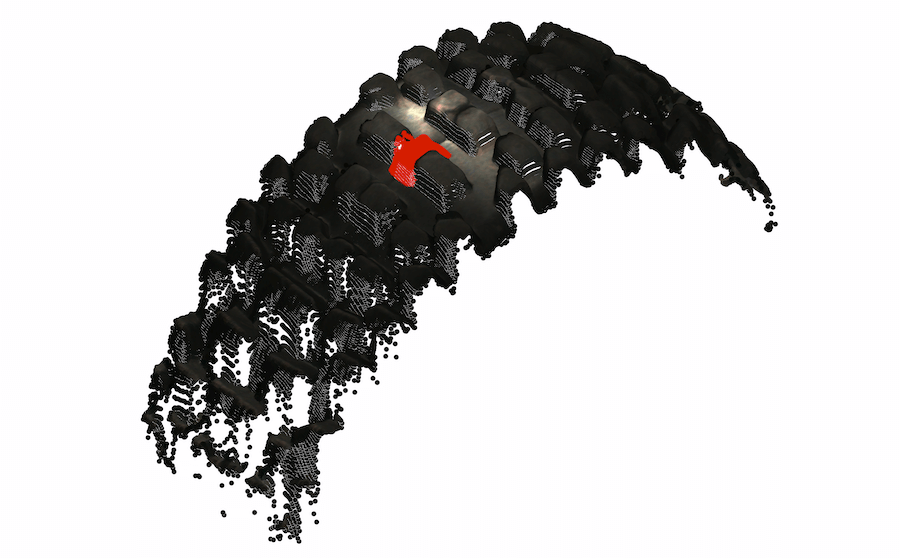}}\\
\rotatebox[origin=c]{90}{Output} &
\raisebox{-0.5\height}{\includegraphics[width=0.25\linewidth]{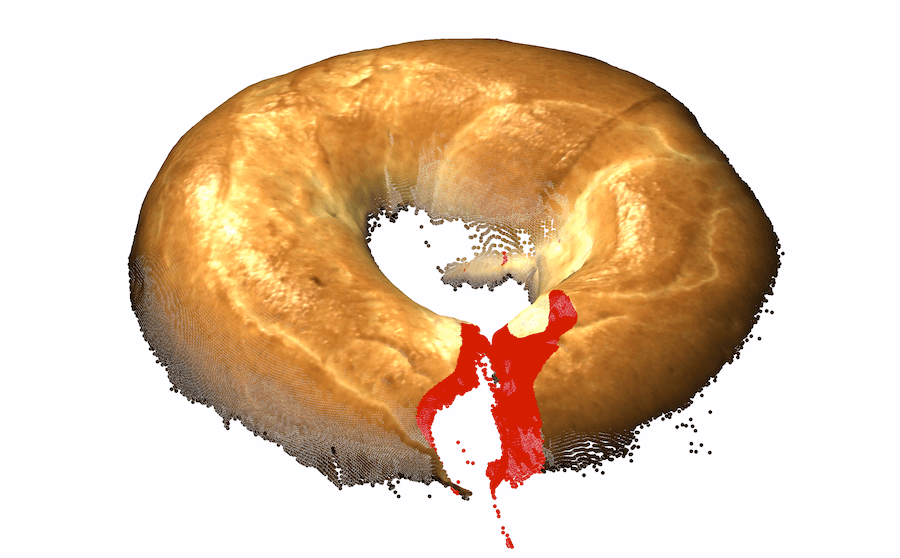}} & 
\raisebox{-0.5\height}{\includegraphics[width=0.25\linewidth]{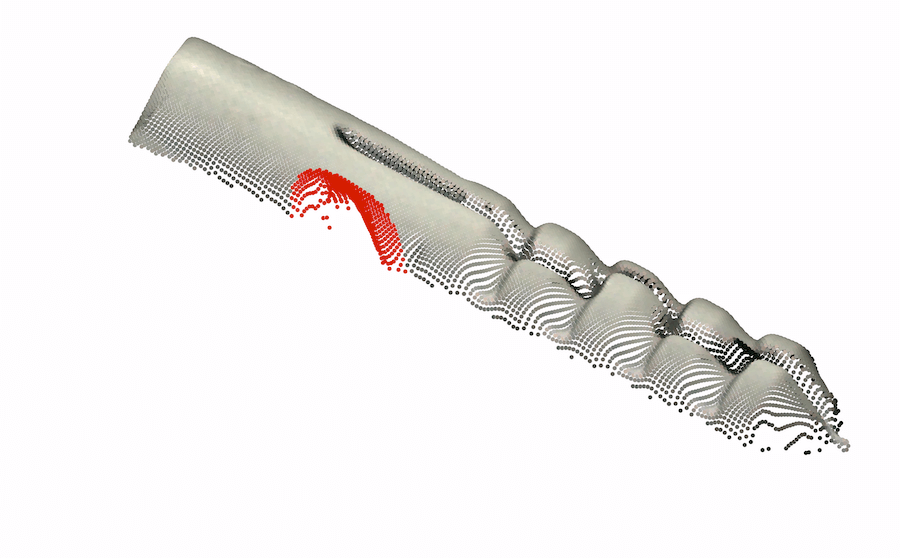}} & 
\raisebox{-0.5\height}{\includegraphics[width=0.25\linewidth]{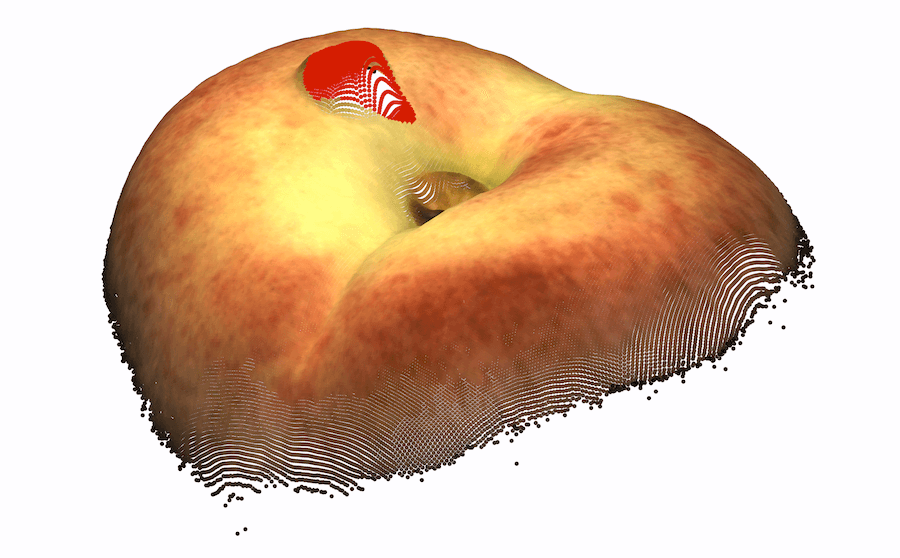}} & 
\raisebox{-0.5\height}{\includegraphics[width=0.25\linewidth]{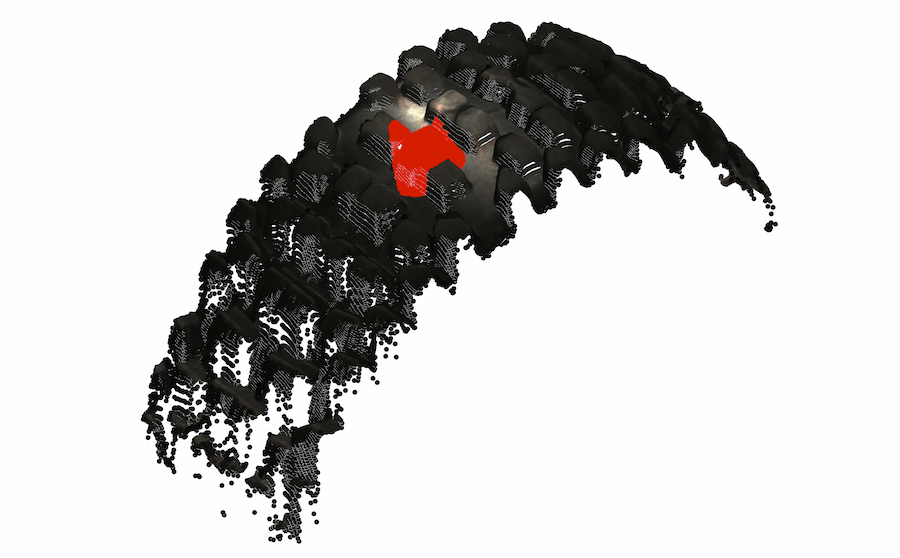}}\\
\end{tabular}
 \caption{\textit{\textbf{Additional results of our method (\textit{BTF}):}} All anomalies are correctly segmented (marked in red)}
\label{fig:more}
\end{figure*}

\section{Related Work}
\textbf{Anomaly detection and segmentation.} Anomaly detection methods have been researched for several decades, most approaches are based either on density estimation or out-of-domain generalization ideas. Classical approaches include: k-Nearest-Neighbors (kNN) \cite{eskin2002geometric}, KDE \cite{latecki2007outlier}, GMM \cite{glodek2013ensemble}, PCA \cite{jolliffe2011principal}, one class SVM (OCSVM) \cite{scholkopf2000support}, and isolation forests \cite{liu2008isolation}. With the advent of deep learning, these methods were extended with deep representations including: DAGMM \cite{zong2018deep} extending PCA, and DeepSVDD \cite{ruff2018deep} extending OCSVM. A novel line of work extends self-supervised approaches to anomaly detection, including Golan and El-Yaniv \cite{golan2018deep}, and Hendrycks et al. \cite{hendrycks2019using} that extend RotNet \cite{gidaris2018unsupervised} and CSI \cite{tack2020csi} who extend contrastive methods \cite{he2019moco,he2020momentum,chen2020simple}.  We follow another line of works that assumes the availability of pre-trained representations and combines them with a kNN scoring function. Such works include Perera and Patel \cite{perera2019learning}, and PANDA \cite{panda}. These works have been extended to anomaly segmentation including SPADE \cite{spade}, PADIM \cite{defard2021padim} and PatchCore \cite{patchcore}. Very recent works have used more advanced density estimation models on the extracted representation, an example is FastFlow \cite{yu2021fastflow}. Other approaches for anomaly segmentation include Student-Teacher autoencoder approaches \cite{bergmann2020uninformed} as well as self-supervised methods that synthesize anomalies such as CutPaste \cite{cutpaste} and NSA \cite{nsa}. 

\textbf{Anomaly detection and segmentation with 3D information.} In contrast to the large amount of research on 2D anomaly detection approaches, 3D anomaly detection has not been extensively researched. In medical imaging research, work was performed to adapt anomaly detection methods to voxel data. Simarro et al. \cite{simarro2020unsupervised} extend f-Anogan \cite{schlegl2017unsupervised,schlegl2019f} to 3D. Bengs et al. \cite{bengs2021three} presented a 3D autoencoder approach for medical voxel data. Voxel data is significantly different from point cloud 3D data. Bergmann et al. \cite{mvtec3d} recognized that a dataset for anomaly segmentation in 3D point cloud data is missing and introduced MVTec 3D-AD\cite{mvtec3d}. We expect this to be a critical contribution to the development of 3D anomaly detection and segmentation. Concurrently to our work, Bergmann and Sattlegger \cite{bergmann2022anomaly} introduced a 3D point cloud based approach dubbed $3D-ST_{128}$ for anomaly detection, we include this work in our investigation.

\section{Problem Definition}
\subsection{Setting}
We assume a set of input training samples $x_1,x_2..x_N$ that are all normal. At test time, we are given a test sample $y$. The goal of anomaly detection is to learn a sample-level scoring function $\sigma_d$, such that $\sigma_d(y)~>~0$ for anomalous samples and $\sigma_d(y)~\leq~0$ for normal ones. The goal of anomaly segmentation is to learn a pixel-level scoring function $\sigma_s$, which satisfies $\sigma_s(y, i)~>~0$ if pixel $i$ of sample $y$ is anomalous, and $\sigma_s(y, i)~\leq~0$ if it is normal.

Many current state-of-the-art methods (e.g. SPADE \cite{spade}, PatchCore \cite{patchcore}) follow the following stages: i) Extracting representation of local regions ii) Estimating the probability density of normal local regions. For example, PatchCore and SPADE perform the density estimation by the nearest-neighbor distance to the normal training dataset. 

\textbf{Representation.} We first compute a representation of each local region that may consist of one or more pixels. The representation of region $j$ of image $x$ is denoted $\phi(x, j)$. In this paper we focus on the representation stage, particularly, our goal is to find learned or handcrafted representations for 3D AD\&S.

\textbf{Anomaly scoring.} Given the representations for every local region $j$ of every training image $x$, we can train a model $\sigma_s(y, j)$ which computes the likelihood of a new representation $\phi(y, j')$. Although some approaches train parametric models for the density of the representations, non-parametric approaches are much simpler and require no training. Specifically, we use the k-Nearest-Neighbor distance of representation $\phi(y, j')$ to the set of all training representations $S = \{\phi(x, j)~~\forall x~~\forall j\}$. Despite their simplicity, such approaches are very accurate, require no training, and can be significantly sped up.

\subsection{3D Representations}
Although RGB images are the default modality, they lack explicit 3D information. Other representations contain direct 3D information e.g., depth maps, \textit{organized} point clouds, \textit{unorganized} point clouds, and voxels. Both organized and unorganized point clouds represent the XYZ location of points in 3D space. However, Organized point clouds retain spatial relation and can thus be treated as images, allowing the use of RGB-based methods (e.g. CNN). Unorganized point clouds, in contrast, do not retain spatial relation and thus require specific methods and models. Finally, voxels are derived from point clouds and can be thought of as a 3D extension of pixels. For brevity we use the term ``pixel`` throughout the paper, however, depending on the context, it may refer to any of the above representations.

\begin{figure*}[t]
\begin{tabular}{@{\hskip1pt}c@{\hskip1pt}c@{\hskip1pt}c@{\hskip1pt}c@{\hskip1pt}c}
3D Input & 2D View & GT & RGB iNet & Depth iNet\\
\includegraphics[width=0.20\linewidth]{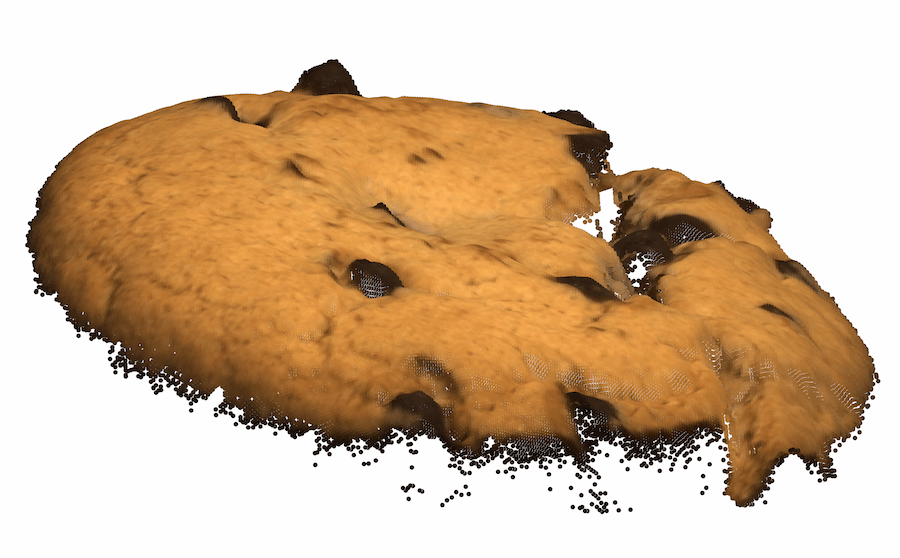} & 
\includegraphics[width=0.20\linewidth]{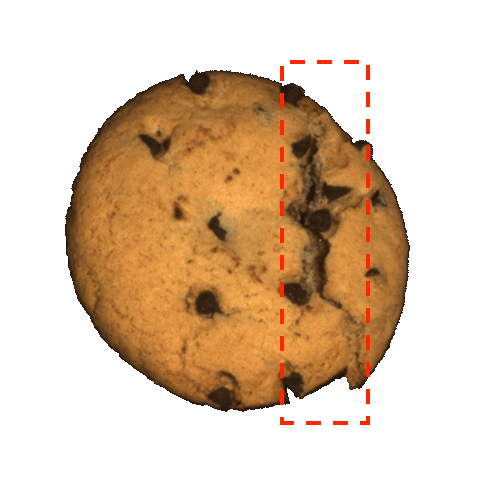} & 
\includegraphics[width=0.20\linewidth]{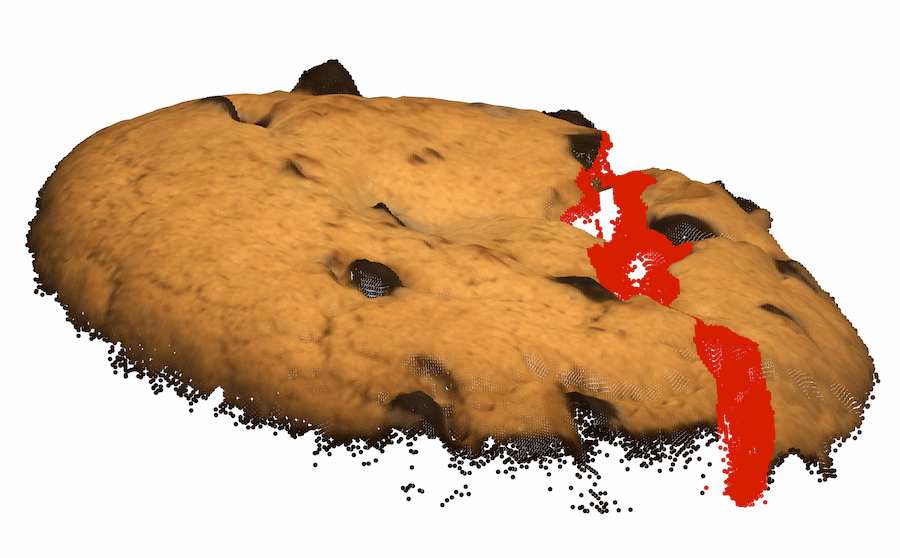} & 
\includegraphics[width=0.20\linewidth]{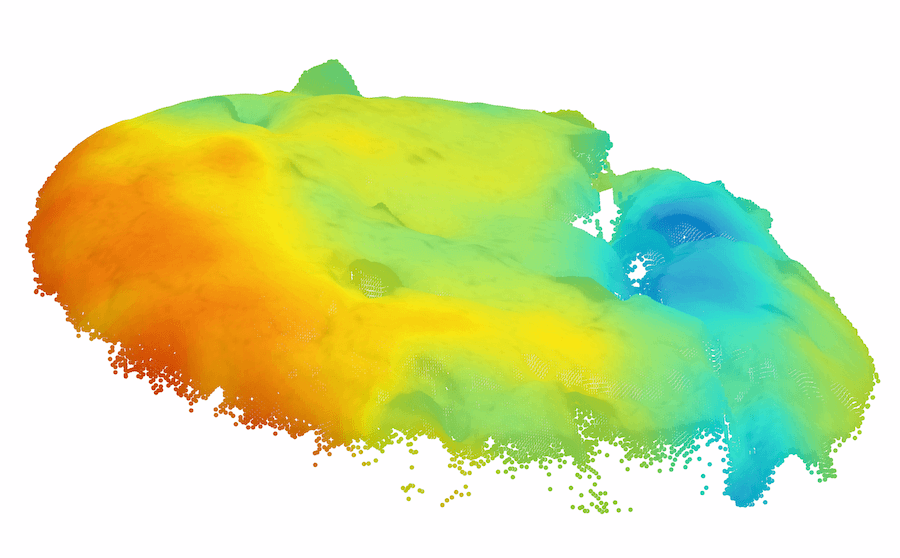} & 
\includegraphics[width=0.20\linewidth]{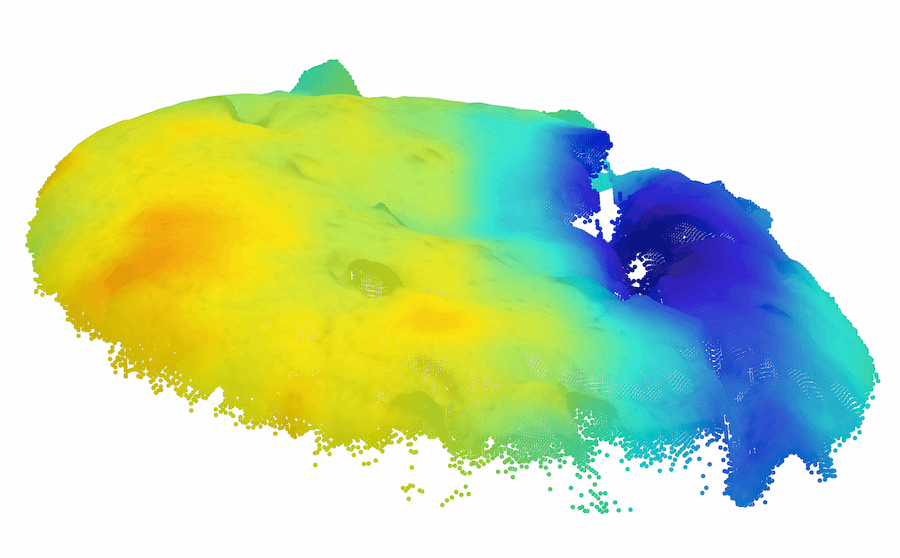} \\
\includegraphics[width=0.20\linewidth]{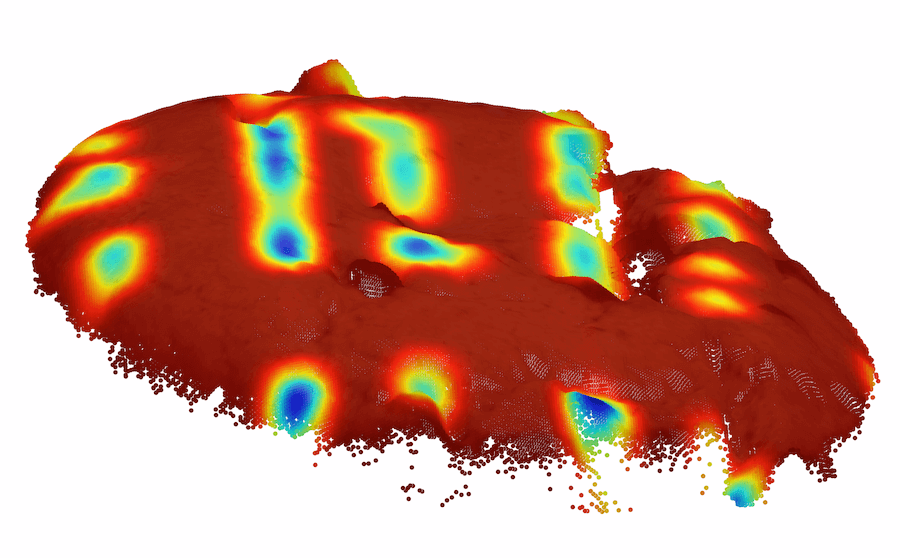} & 
\includegraphics[width=0.20\linewidth]{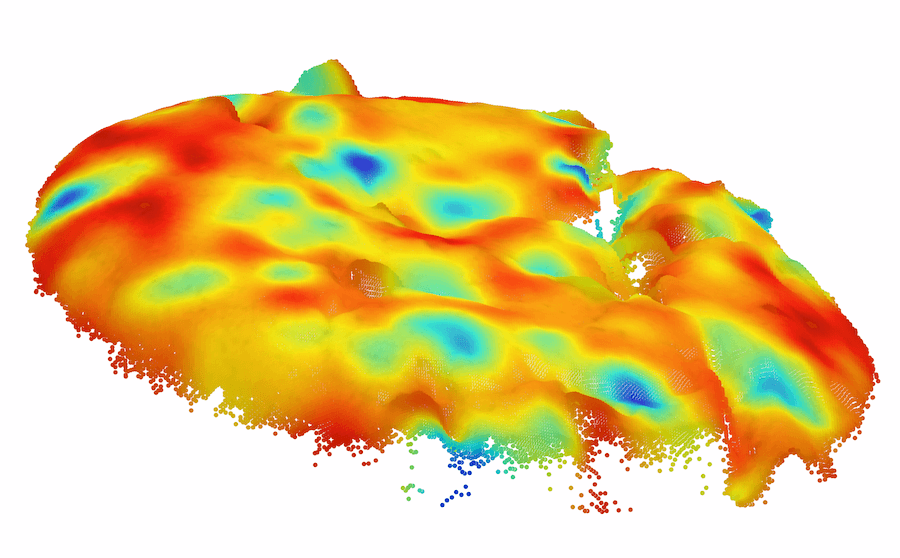} & 
\includegraphics[width=0.20\linewidth]{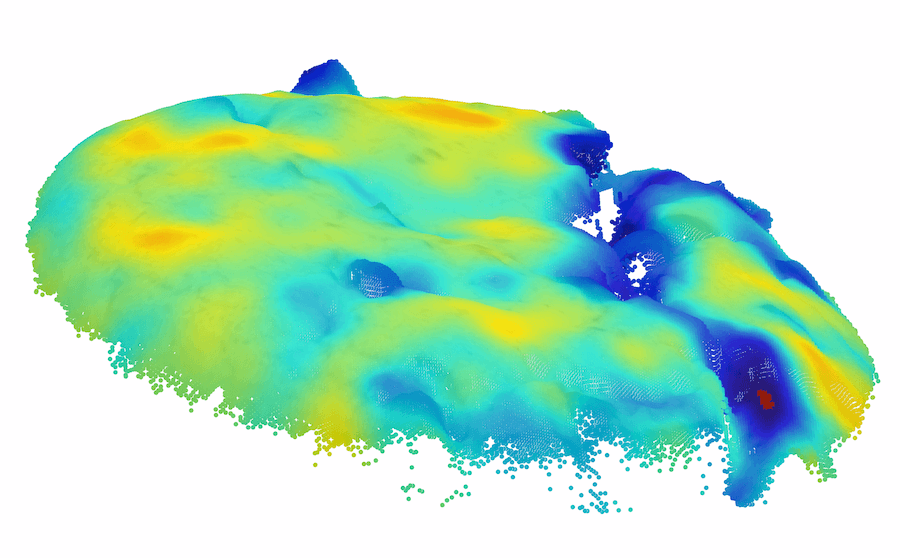} & 
\includegraphics[width=0.20\linewidth]{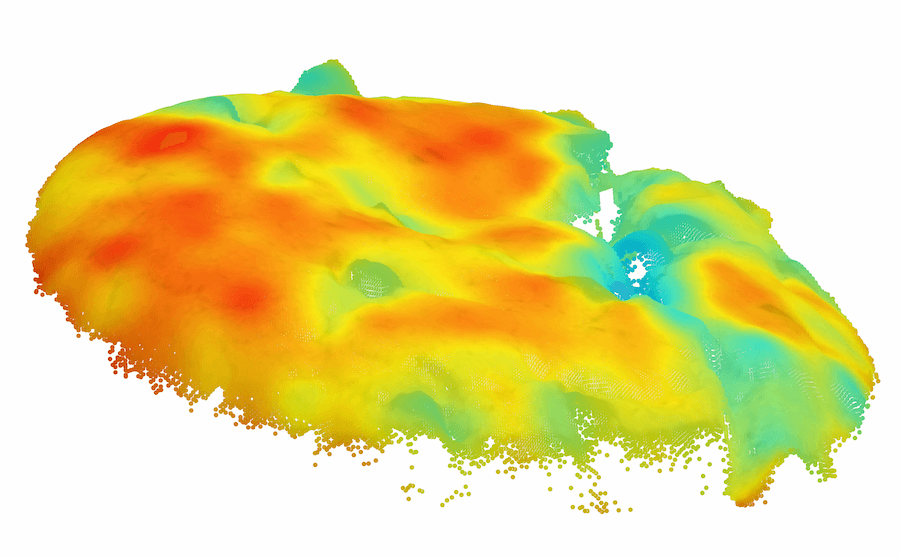} & 
\includegraphics[width=0.20\linewidth]{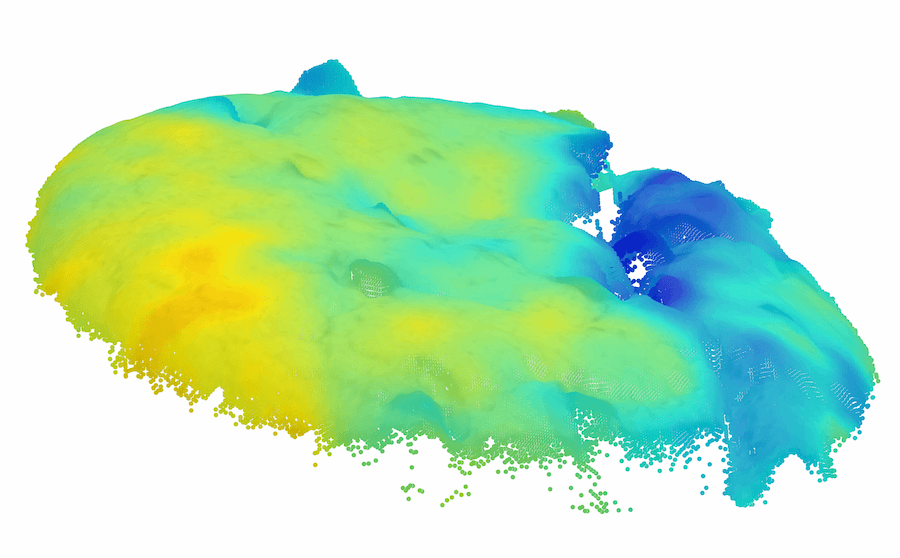} \\
Raw & HoG & SIFT & FPFH & BTF \\
\end{tabular}
 \caption{\textit{\textbf{Anomaly Heatmaps (Pixel-wise):}} The distance heatmaps are shown for each method. The \textit{PRO} and \textit{P-ROC} metrics use these distances for the final score. Blue colors indicate large distances (i.e., more anomalous) while red colors are smaller distances (i.e., less anomalous). Anomalies are indicated in red. ``iNet`` indicates ImageNet pre-trained features}
\label{fig:heatmaps}
\end{figure*}

\subsection{Benchmark}
Our investigation uses the recently published MVTec 3D Anomaly Detection dataset \cite{mvtec3d}. It contains over $4000$ high-resolution 3D scans of industrially manufactured products across $10$ categories. Each sample is represented by an \textit{organized} point cloud and a corresponding RGB image with a one-to-one mapping between the pixels in the point cloud and those in the RGB image. Five of the classes in the dataset exhibit natural variations (\emph{bagel}, \emph{carrot}, \emph{cookie}, \emph{peach}, and \emph{potato}). The classes \emph{cable gland} and \emph{dowel} are of rigid bodies, while the classes \emph{foam}, \emph{rope}, and \emph{tire} are ``man-made`` but deformable. Bergmann et al. \cite{mvtec3d} introduced three baselines for the dataset: GAN-based, Autoencoder-based (AE), and Variation Model (VM) - a simple baseline based on per-pixel mean and standard deviation. These models operate either on the depth images or in voxel space with additional variants that operate on 3D+RGB information.

\subsection{Evaluation Metrics}
We use several evaluation metrics. Image-level anomaly detection is measured using image-level ROCAUC \cite{rocauc}  (denoted \textit{I-ROC}). Two pixel-level metrics are used for anomaly segmentation: i) pixel-wise ROCAUC, an extension of the standard ROCAUC for the pixel level which simply treats each pixel in the dataset as a sample and computes the ROCAUC over all pixels in the dataset (denoted \textit{P-ROC}). ii) The \textit{PRO} \cite{mvtec2d} metric, defined as the average relative overlap of the binary prediction $P$ with each ground truth connected component $C_k$ where $K$ denotes the number of ground truth components. The final metric is computed by integrating this curve up to some false positive rate and normalizing

\begin{equation*}
    \mathrm{PRO} = \frac{1}{K} \sum_{k=1}^{K} \frac{|P \cap C_{k}|}{|C_{k}|},
\end{equation*}

 Following common practice, unless otherwise stated, we use the integration limit of $0.3$.


\section{An Empirical Investigation of 3D AD\&S}
\subsection{Do current 3D methods beat 2D methods?}
\label{sec:2d_v_3d}

    

We begin our investigation by evaluating if current 3D AD\&S methods are actually better than the SoTA 2D methods when applied on \textbf{3D} data. To represent 3D methods, we test two approaches: i) Voxel GAN \cite{mvtec3d}, a generative method proposed as a baseline for 3D AD\&S. While it has several variants, we use the best performing ones, which are ``Voxel`` and ``Voxel + RGB``. ii) 3D-ST \cite{3d_ad_pc_deep}, a concurrent method that uses a point cloud student-teacher model to learn 3D representations. We use PatchCore \cite{patchcore} to represent color-based image AD\&S methods. Importantly, PatchCore uses features that were pre-trained on the ImageNet \cite{imagenet} dataset, which has been shown to be highly effective for image AD\&S. In contrast, 3D-ST used ModelNet10 \cite{modelnet10} for pre-training their teacher model. We present the results in Fig.~\ref{table:2d_v_3d}. Surprisingly, PatchCore, which does not use 3D information, outperforms all previous methods.

\textbf{Conclusion.} Currently, state-of-the-art methods for image AD\&S that use only color information, outperform 3D AD\&S methods that use 3D or 3D + color information.\\

\subsection{Is 3D information potentially useful for AD\&S?}
\label{sec:3d_util}
Provided the results of Sec.~\ref{sec:2d_v_3d}, we are faced with a second question: ``Is 3D information \textit{potentially} useful for AD\&S$?$``. Below we present two cases in which 3D information is indeed useful for AD\&S.


\textit{Ambiguous geometry.} Frequently, we are unable to determine the underlying geometry of an object by only looking at the color information of the object. In such cases, 3D information may reveal the true geometry. We present several examples of such cases in the left half of Fig.~\ref{fig:rgb_v_d}-top~row, the anomaly in each object cannot be detected from color information only. In the bottom row, using the 3D information, we present another view of the same objects where the anomalies are easily detected. In the case of the cookie\footnotemark[1], looking at the color-only image, the hole blends in with the rest of the chocolate chips, making it hard to visually identify the image as anomalous. 
Using the 3D information, we visualize the cookie from a different angle, making it easy to spot the anomaly. Looking at the image of the potato, it is hard to infer the geometry of the dent from shadow and texture. However, viewing the potato from different angles (by using the 3D information), the different texture reveals the dent. 

\begin{figure}[t]
\begin{subtable}{0.5\textwidth}
\begin{tabular}{c@{\hskip5pt}c@{\hskip5pt}c@{\hskip5pt}c@{\hskip5pt}c@{\hskip5pt}c@{\hskip5pt}c@{\hskip5pt}c@{\hskip5pt}} 
    \toprule
    \multicolumn{2}{c}{Voxel} & \multicolumn{2}{c}{Voxel + RGB} & \multicolumn{2}{c}{Point Cloud} & \multicolumn{2}{c}{RGB}\\
    \multicolumn{2}{c}{GAN} & \multicolumn{2}{c}{GAN} & \multicolumn{2}{c}{$3D-ST_{128}$} & \multicolumn{2}{c}{PatchCore}\\
    \cmidrule(r){1-2}
    \cmidrule(r){3-4}
    \cmidrule(r){5-6}
    \cmidrule(r){7-8}
    \textit{PRO} & \textit{I-ROC} & \textit{PRO} & \textit{I-ROC} & \textit{PRO} & \textit{I-ROC} & \textit{PRO} & \textit{I-ROC}\\
    
    \cmidrule(r){1-1}
    \cmidrule(r){2-2}
    \cmidrule(r){3-3}
    \cmidrule(r){4-4}
    \cmidrule(r){5-5}
    \cmidrule(r){6-6}
    \cmidrule(r){7-7}
    \cmidrule(r){8-8}
    0.583 & 0.537 & 0.639 & 0.517 & 0.833 & - & \textbf{0.876} & \textbf{0.785}\\
   \bottomrule
    \end{tabular}
    \caption{\textit{\textbf{Current 3D methods vs. SoTA 2D method}}}
    \label{table:2d_v_3d}
\end{subtable}
\begin{subfigure}{0.5\textwidth}
    \centering
    \includegraphics[width=0.7\linewidth]{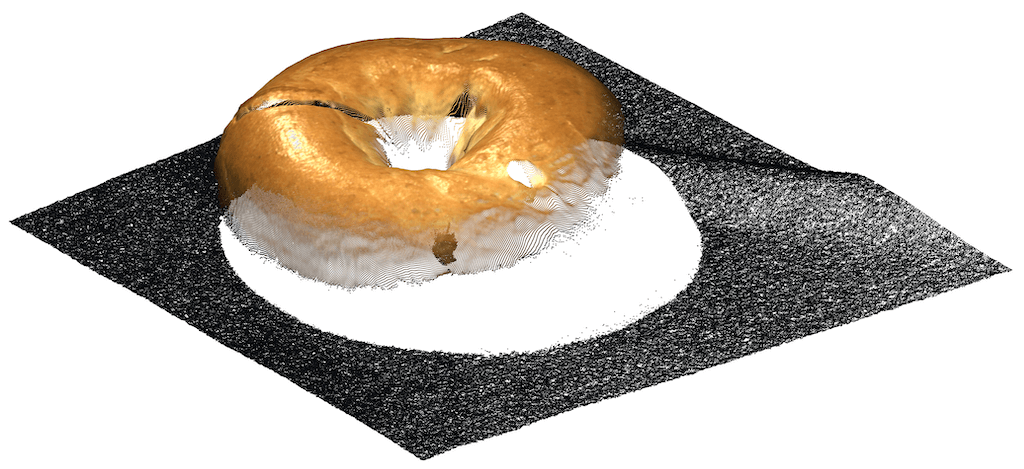}
    \caption{\textit{\textbf{3D-Aware Preprocessing}}}
\label{fig:artifact}
\end{subfigure}
\caption{\textit{\textbf{Current 3D methods vs. SoTA 2D method (left):}} We compare recent leading 3D-based methods against the current SoTA 2D AD\&S method (PatchCore), surprisingly, PatchCore, which does not use 3D information, outperforms all baselines. Average metrics across all classes, best performing MVTec 3D-AD methods are shown.\\ \textit{\textbf{3D-Aware Preprocessing (right):}} A nuisance artifact in the fabric}
\end{figure}

\textit{Background variation.} 
Curated datasets usually contain synthetic conditions such as centered objects and clean backgrounds, but the reality is seldom so simple. Many methods mistakenly classify cluttered image backgrounds as anomalous. Although background segmentation is not trivial, it is far easier when the 3D information is provided. We found cases in the MVTec-3D datasets where  backgrounds nuisance artifacts triggered false-positive alerts. We demonstrate such a case in Fig.~\ref{fig:artifact}, the background fabric contains ``wave`` like patterns which are hard to detect given the dark background color.

\textbf{Conclusion.} 3D information is often required to identify anomalies, even when color is available.

\subsection{What are the key properties of successful 3D AD\&S representations?}
\label{sec:3d_rep_study}
Having shown that 3D information is under-utilized by current methods, and having established the necessity of 3D information for image AD\&S, we now seek to answer a third question: ``What are the key properties of successful 3D AD\&S representations$?$``. We distinguish among several categories.\\

\noindent \textbf{Learning-based representations designed for images.} We adapt the two most popular learning-based image AD\&S paradigms to 3D data: i) ImageNet pre-trained features ii) Self-supervised methods. 

\noindent  \textit{Depth-only ImageNet features.} Motivated by the impressive results of ImageNet pre-trained features on color images (Sec.~\ref{sec:2d_v_3d}), we apply PatchCore on depth images.

\noindent   \textit{NSA.} A different class of learning-based methods approaches AD\&S from a generative perspective. CutPaste and NSA \cite{cutpaste,nsa} are recent works that try to mimic anomalies by pasting image patches at different image locations. Specifically, NSA uses Poisson blending \cite{poisson_blending} to make these augmentations appear more natural.

\noindent  \textit{Results.} ImageNet pre-trained features significantly outperform NSA on depth images (Tab.~\ref{table:3d_rep_summ}). Both approaches underperform PatchCore applied to color images.

\noindent \textbf{Handcrafted Image Representations.} Depth patterns are often much simpler than color patterns. We hypothesize that a simple, handcrafted descriptor should suffice. The following depth representations do not require external data or training.

\noindent  \textit{Raw Depth Values.} Here, we test perhaps the simplest possible representation, the raw depth values of a patch.

\noindent  \textit{Histogram of Oriented Gradients (HoG).} HoG \cite{hog} considers image gradients and uses histograms to capture the distribution of gradient orientations in a patch. This is potentially more powerful than raw values as the descriptor encodes the spatial structure of the data while being invariant to small translations. On the other hand, HoG is not invariant to global rotations, a much-desired property for 3D representations. Additionally, the small context of HoG makes it invariant to local geometric changes. This is counterproductive to our goal of detecting anomalies - usually manifested as local geometric changes.

\noindent  \textit{Dense Scale-Invariant Feature Transform (D-SIFT).} In contrast to HoG, SIFT \cite{sift} is rotation, scale, and shift-invariant as it is rotated to align the most dominant direction to the base orientation. This reduces the rotation ambiguity allowing matches between rotated images. 

\noindent \textit{Results.} HoG significantly boosts pixel-level accuracy, achieving better results than raw and learning-based features. These strong results are obtained despite HoG not being specifically designed for 3D information. Finally, the D-SIFT descriptor is able to surpass all previous depth-based results (including learning-based ones) on all three metrics.

\noindent \textbf{3D rotation-Invariant Representations.} Rotation-invariant features were very effective on depth maps. We now ask if rotation-invariant 3D features can do better?

\noindent \textit{Fast Point Feature Histograms (FPFH) \cite{fpfh}.} The method first computes the k-Nearest-Neighboring points to the region center point. It then computes a histogram-based representation as a function of the surface normals and vector distance to the nearest neighbors. We choose this as the representative due to its time-tested excellent performance.

\begin{table*}[t]
\begin{center}
\caption{\textit{\textbf{Summary of Our Findings:}} Average metrics across all classes, ``iNet`` indicates ImageNet pre-trained, PC indicates point cloud}
\label{table:3d_rep_summ}

\begin{tabular}{l@{\hskip5pt}|c@{\hskip5pt}c@{\hskip5pt}c@{\hskip5pt}c@{\hskip5pt}c@{\hskip5pt}c@{\hskip5pt}c@{\hskip5pt}c@{\hskip5pt}c@{\hskip5pt}c@{\hskip5pt}} 
    \toprule
    Modality & RGB & Depth & Depth & Depth & Depth & Depth & PC &  PC + RGB & PC & RGB + PC \\
    Method & iNet & iNet & NSA & Raw & HoG & SIFT & FPFH  & PointNeXt & SpinNet & BTF\\
    \midrule
    PRO & 0.876 & 0.586 & 0.572 & 0.191 & 0.614 & 0.866 & 0.924 &  0.380 & 0.654 &\textbf{0.964}\\
    I-ROC & 0.785 & 0.637 & 0.696 & 0.528 & 0.560 & 0.714 & 0.753 & 0.587 & 0.524 & \textbf{0.865}\\
    P-ROC & 0.966 & 0.821 & 0.817 & 0.548 & 0.845 & 0.954 & 0.980 &  0.687 & 0.873 &\textbf{0.993}\\
   \bottomrule
   
    \end{tabular}
    
\end{center}
\end{table*}

\begin{table*}[t]
\begin{center}
\caption{\textbf{\textit{Detailed PRO Results:}} Top half are current state-of-the-art, bottom half are methods investigated by us. Many of our methods outperform all current methods by a wide margin. ``iNet`` indicates ImageNet pre-trained}
\label{table:au_pro}
\resizebox{\linewidth}{!}{%
\begin{tabular}{@{\hskip5pt}l@{\hskip5pt}l|cccccccccc|c} 
    \toprule
    & Method & Bagel & \begin{tabular}[c]{@{}c@{}}Cable\\ Gland\end{tabular} & Carrot & Cookie & Dowel & Foam & Peach & Potato & Rope & Tire & Mean \\ 
    \midrule

    \multirow{13}{*}{\rotatebox[origin=c]{90}{Previous Methods}} & Voxel GAN & 0.440 & 0.453  & 0.825 & 0.755 & 0.782 & 0.378 & 0.392 & 0.639 & 0.775 & 0.389 & 0.583 \\
    & \qquad \qquad  + RGB & 0.664 & 0.620  & 0.766 & 0.740 & 0.783 & 0.332 & 0.582 & 0.790 & 0.633 & 0.483 & 0.639 \\
    & Voxel AE & 0.260 & 0.341  & 0.581 & 0.351 & 0.502 & 0.234 & 0.351 & 0.658 & 0.015 & 0.185 & 0.348 \\
    & \qquad \qquad  + RGB & 0.467 & 0.750  & 0.808 & 0.550 & 0.765 & 0.473 & 0.721 & 0.918 & 0.019 & 0.170 & 0.564 \\
    & Voxel VM & 0.453 & 0.343  & 0.521 & 0.697 & 0.680 & 0.284 & 0.349 & 0.634 & 0.616 & 0.346 & 0.492 \\ 
    & \qquad \qquad  + RGB & 0.510 & 0.331  & 0.413 & 0.715 & 0.680 & 0.279 & 0.300 & 0.507 & 0.611 & 0.366 & 0.471 \\
    \cmidrule{2-13}
    & Depth GAN & 0.111 & 0.072  & 0.212 & 0.174 & 0.160 & 0.128 & 0.003 & 0.042 & 0.446 & 0.075 & 0.143 \\
    & \qquad \qquad  + RGB & 0.421 & 0.422  & 0.778 & 0.696 & 0.494 & 0.252 & 0.285 & 0.362 & 0.402 & 0.631 & 0.474 \\
    & Depth AE & 0.147 & 0.069  & 0.293 & 0.217 & 0.207 & 0.181 & 0.164 & 0.066 & 0.545 & 0.142 & 0.203 \\
    & \qquad \qquad  + RGB & 0.432 & 0.158  & 0.808 & 0.491 & 0.841 & 0.406 & 0.262 & 0.216 & 0.716 & 0.478 & 0.481 \\
    & Depth VM &  0.280 & 0.374  & 0.243 & 0.526 & 0.485 & 0.314 & 0.199 & 0.388 & 0.543 & 0.385 & 0.374 \\
    & \qquad \qquad  + RGB & 0.388 & 0.321  & 0.194 & 0.570 & 0.408 & 0.282 & 0.244 & 0.349 & 0.268 & 0.331 & 0.335 \\
    \cmidrule{2-13}
    & $3D-ST_{128}$ & 0.950 & 0.483 & \textbf{0.986} & 0.921 & 0.905 & 0.632 & 0.945 & \textbf{0.988} & 0.976 & 0.542 & 0.833 \\
    \midrule
    \multirow{10}{*}{\rotatebox[origin=c]{90}{Our Findings}} & RGB iNet & 0.898 & 0.948 & 0.927 & 0.872 & 0.927 & 0.555 & 0.902 & 0.931 & 0.903 & 0.899 & 0.876 \\
    & Depth iNet & 0.701 & 0.544 & 0.791 & 0.835 & 0.531 & 0.100 & 0.800 & 0.549 & 0.827 & 0.185 & 0.586 \\
    & NSA & 0.724  &  0.228 &  0.716 & 0.856  & 0.320 & 0.432 & 0.712 & 0.655 & 0.818 & 0.258 & 0.572 \\
    & Raw & 0.040 & 0.047 & 0.433 & 0.080 & 0.283 & 0.099 & 0.035 & 0.168 & 0.631 & 0.093 & 0.191 \\
    & HoG & 0.518 & 0.609 & 0.857 & 0.342 & 0.667 & 0.340 & 0.476 & 0.893 & 0.700 & 0.739 & 0.614 \\
    & SIFT & 0.894 & 0.722 & 0.963 & 0.871 & 0.926 & 0.613 & 0.870 & 0.973 & 0.958 & 0.873 & 0.866 \\
    & FPFH & 0.972 & 0.849 & 0.981 & 0.939 & 0.963 & 0.693 & 0.975 & 0.981 & \textbf{0.980} & 0.949 & 0.928 \\
    & PointNext  & 0.425 & 0.294 & 0.365 & 0.772 & 0.227 & 0.151 & 0.408 & 0.101 & 0.771 & 0.295 & 0.380 \\
    & SpinNet & 0.635 & 0.316 & 0.922 & 0.780 & 0.870 & 0.380 & 0.585 & 0.699 & 0.955 & 0.400 & 0.654 \\
    & BTF & \textbf{0.976} & \textbf{0.967} & 0.979 & \textbf{0.974} & \textbf{0.971} & \textbf{0.884} & \textbf{0.976} & 0.981 & 0.959 & \textbf{0.971} & \textbf{0.964} \\
   \bottomrule
    \end{tabular}
}
\end{center}
\end{table*}

\noindent \textbf{Point-cloud specific learning-based representations.}

\noindent \textit{PointNeXt \cite{point_next}.} A U-Net \cite{unet} architecture in which the encoder hierarchically abstracts the point cloud features while the decoder gradually interpolates the abstracted features. 

\noindent \textit{SpinNet \cite{spinnet}.}  A rotation invariant, learning-based representation learning method. A transformation and voxelization phase make the model rotation invariant.

\noindent \textit{Results.}
Compared to most methods mentioned above, PointNeXt falls short. SpinNet performs better than PointNeXt (another indicator of the importance of the rotation invariance), yet fails to surpass the proposed rotation invariant, handcrafted methods. See Tab.~\ref{table:au_pro}, Fig.~\ref{fig:image_rocauc}, and App.~\ref{app:point_next_results} for the results.

\noindent \textbf{Conclusion.} FPFH outperforms all methods that use color, depth, or both (Tab.~\ref{table:3d_rep_summ}). The results show that strong, handcrafted, rotation-invariant 3D representations are extremely effective for AD\&S when 3D information is available. Furthermore, as anomalies are usually local and ``fine-grained``, using only a small subset of points (as required by many deep-learning-based methods) reduces performance.

\subsection{Are there complimentary benefits from using both 3D and color modalities?}

While the best depth-only representation outperformed existing color-only representations, we hypothesize that combining them might achieve the best of both worlds. In some cases, geometry alone does not suffice for detecting anomalies. Two examples are fine textures and color-based anomalies. The ``cable gland`` in Fig.~\ref{fig:rgb_v_d}-right is slightly scraped. While this anomalous texture is clearly observed in the color image, it is virtually impossible to detect with the current resolution of the 3D information. This is even more apparent in the foam example, wherein the anomaly is manifested as a change in color. As our exclusive focus on 3D fails to account for certain anomalies, it is necessary to combine 3D and color information.

\begin{figure*}[t]
\begin{tabular}{@{\hskip1pt}c@{\hskip1pt}c@{\hskip1pt}c@{\hskip1pt}c@{\hskip1pt}c}
3D Input & 2D View & GT & RGB iNet & Depth iNet\\
\includegraphics[width=0.20\linewidth]{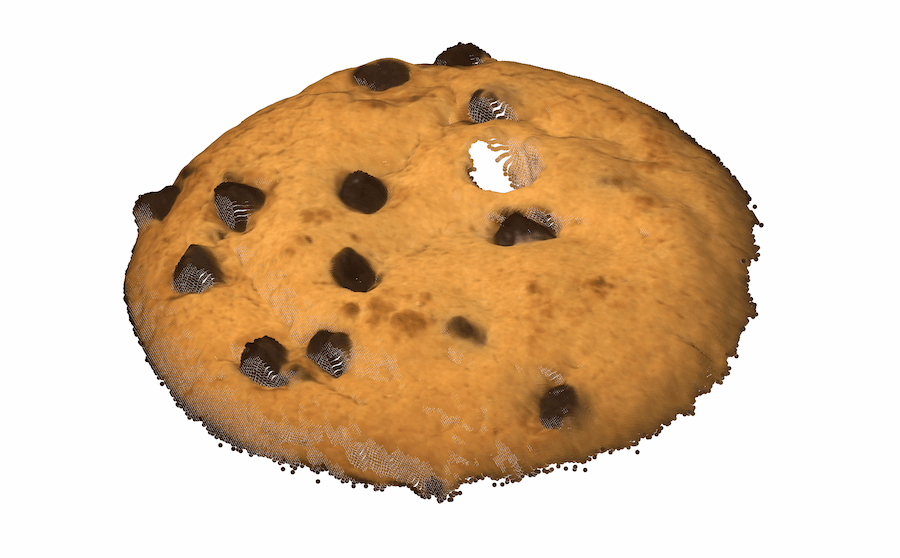} & 
\includegraphics[width=0.20\linewidth]{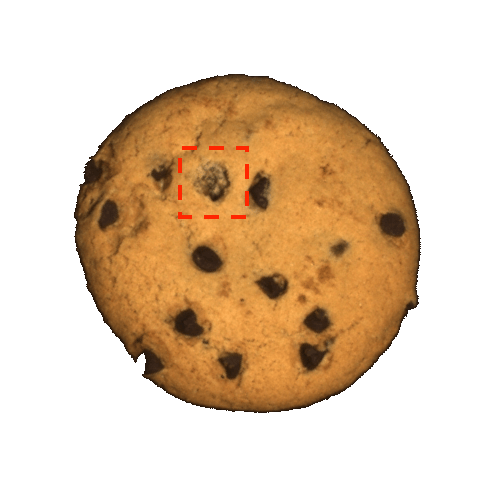} & 
\includegraphics[width=0.20\linewidth]{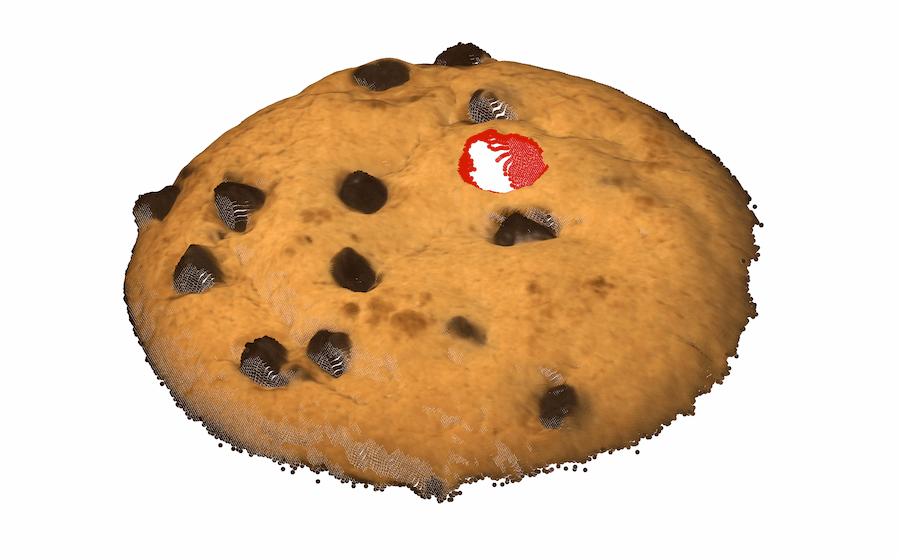} & 
\includegraphics[width=0.20\linewidth]{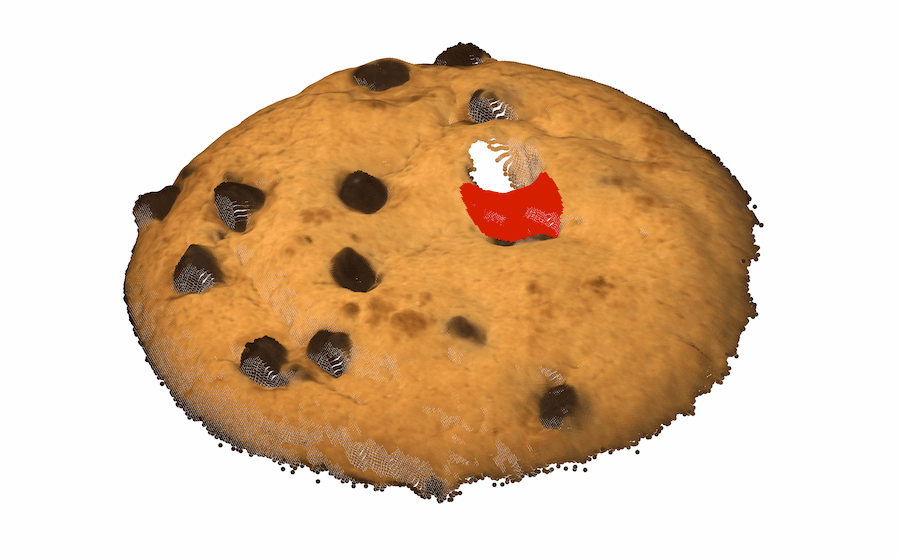} & 
\includegraphics[width=0.20\linewidth]{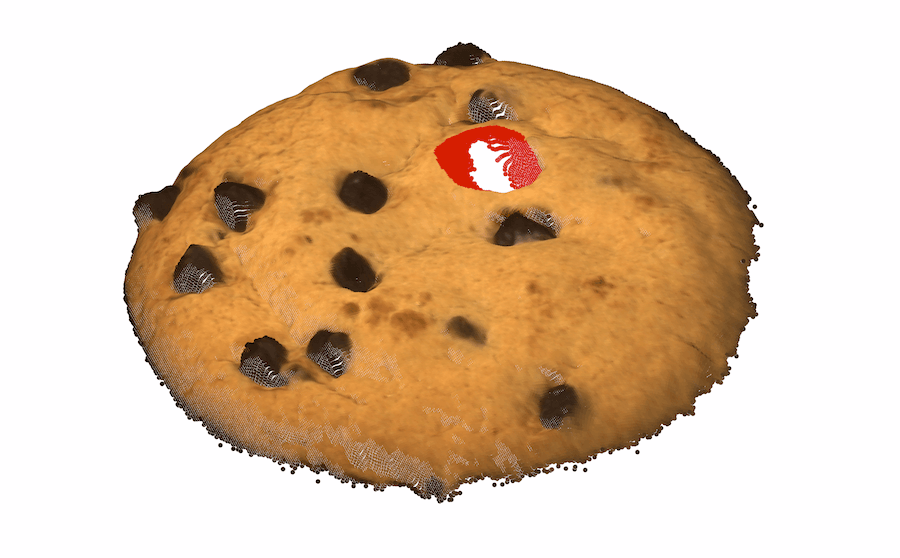} \\
\includegraphics[width=0.20\linewidth]{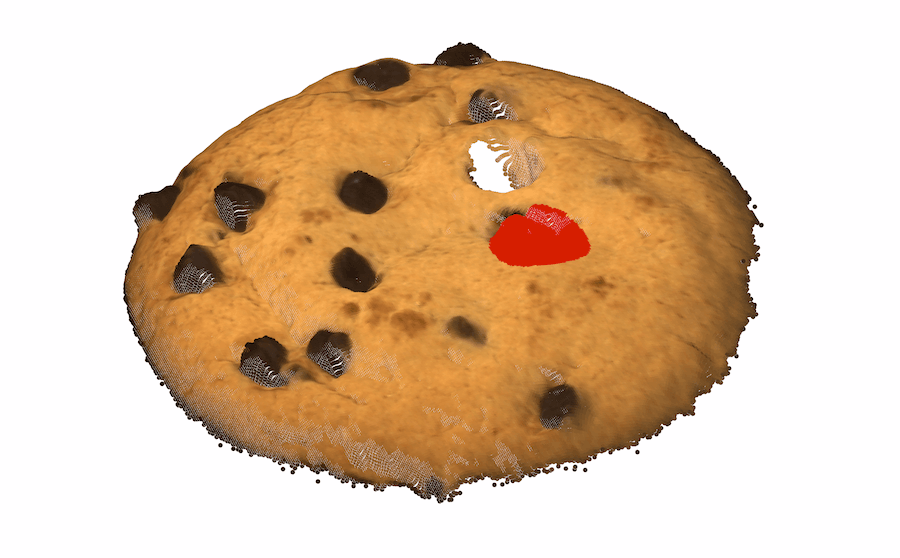} & 
\includegraphics[width=0.20\linewidth]{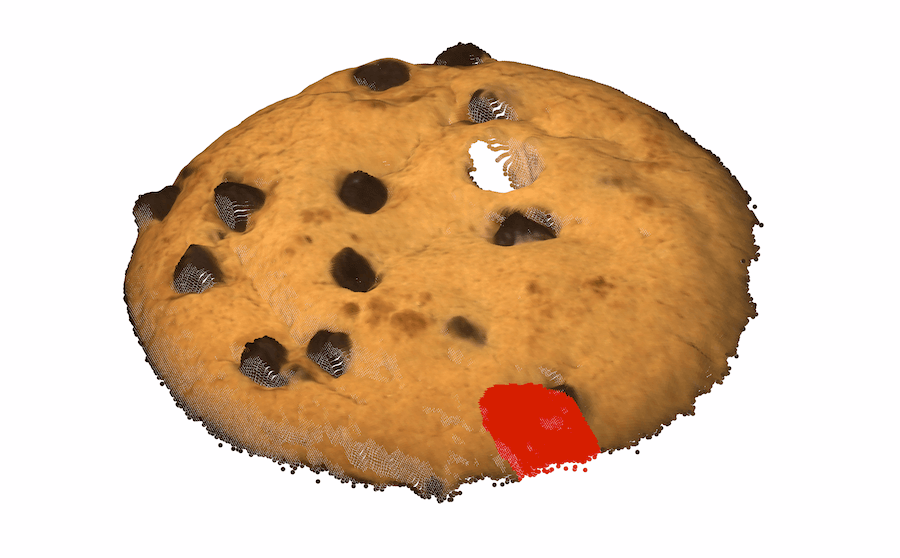} & 
\includegraphics[width=0.20\linewidth]{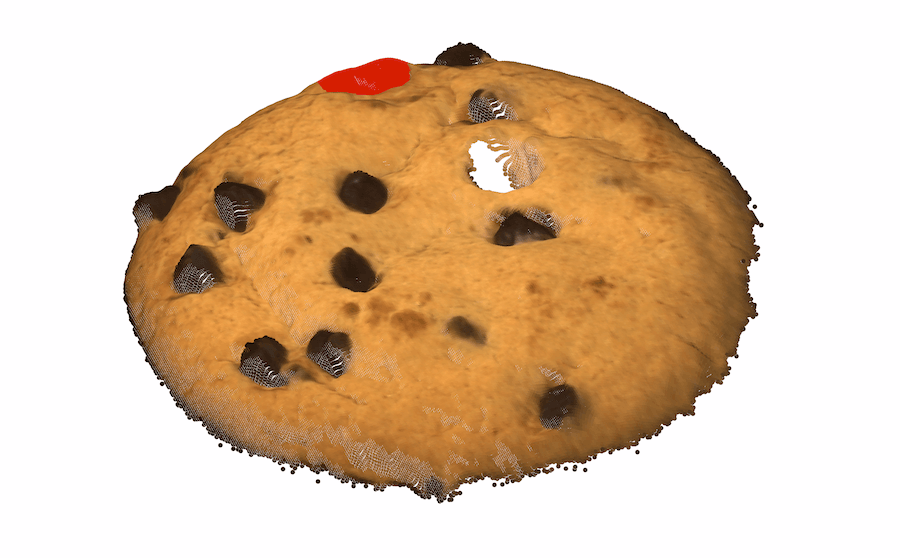} & 
\includegraphics[width=0.20\linewidth]{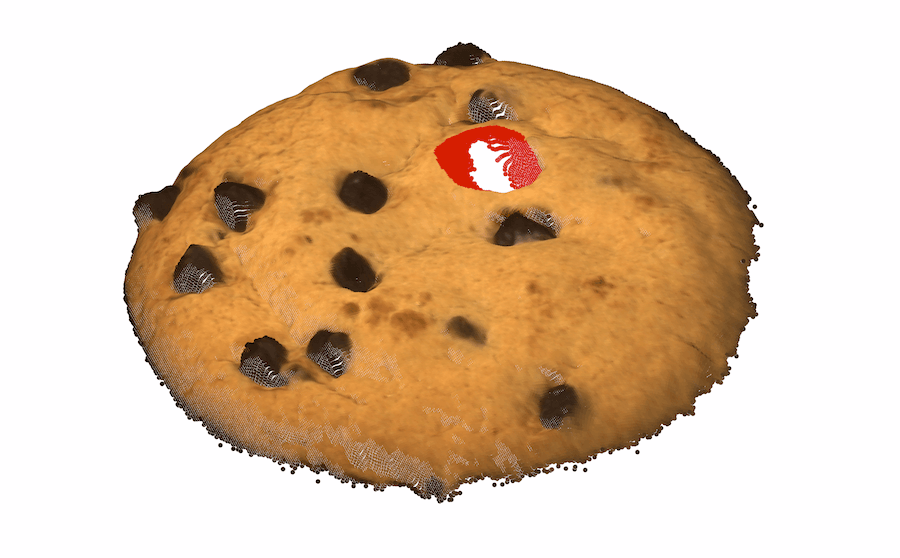} & 
\includegraphics[width=0.20\linewidth]{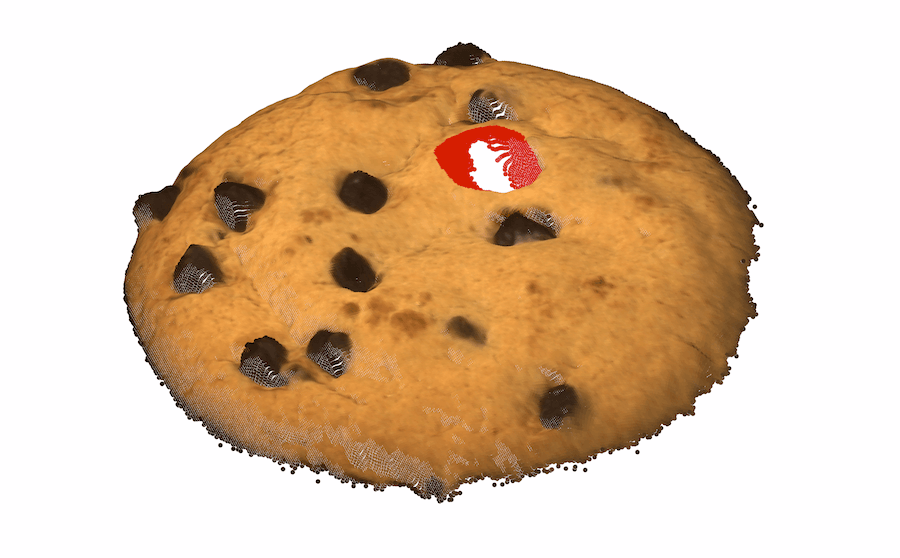} \\
Raw & HoG & SIFT & FPFH & BTF \\
\end{tabular}
 \caption{\textit{\textbf{Most Distant Patch (\textit{I-ROC}):}} The patch with the largest kNN distance is shown in red for each representation. Anomaly indicated by a red square in the 2D view, ``iNet`` indicates ImageNet pre-trained}
\label{fig:max_patch}
\end{figure*}

\textit{BTF - A Combined color + 3D Approach.}
\label{sec:2d_plus_3d}
We take a combined color + 3D approach. To this end, color representations are extracted using the ImageNet-based method discussed in Sec.~\ref{sec:2d_v_3d} and 3D representations are extracted using FPFH as discussed in Sec.~\ref{sec:3d_rep_study}. We concatenate these two representations, forming a color + 3D representation which we dub \textit{BTF} (Back to the Feature).

\noindent \textit{Results.} Compared with the previous best method of combining 3D and RGB (``Voxel GAN + RGB``), our \textit{BTF} improves the \textit{PRO} (i.e. anomaly segmentation) metric by $32.5\%$ and \textit{I-ROC} (i.e. anomaly detection) by $33.6\%$. Compared to using only 3D information, our \textit{BTF} improves on FPFH by $3.6\%$  \textit{PRO} and $12\%$ \textit{I-ROC}. Moreover, it achieves a score of $99.3\%$ on \textit{P-ROC}, a $1.3\%$ improvement over FPFH (Tab.~\ref{table:au_pro}, Fig.~\ref{fig:image_rocauc}). Other color and 3D combinations and extended results are found in App.~\ref{app:rgb_depth_comb} and App.~\ref{app:iroc_results}.

\noindent \textbf{Conclusion.} By combining color and 3D information, our \textit{BTF} representation makes use of complementary attributes from both modalities, achieving the best results to date on the MVTec 3D-AD dataset.


\subsection{Implementation Details}
\label{sec:impl_details}
Unless otherwise stated, the original point clouds and color images are downsampled to $224 \times 224$. For point clouds, we downsample the \textit{organized} point cloud (i.e. image downsampling) using nearest-neighbor interpolation, the color images are downsampled using bicubic interpolation. For \textit{unorganized} point clouds, we reshape the \textit{organized} point cloud from $n\times m \times 3$ into $n \cdot m \times 3$. We use the $Z$ channel of the \textit{organized} point cloud as our depth map. We extract $28\cdot28=784$ patches (features) from each sample, the feature dimension varies based on the representations used. When the representation is extracted at a different resolution, we use average pooling to match $28\cdot28=784$. For nonsquare classes (i.e. \textit{rope} and \textit{tire}), we pad the color and 3D images with zeros. For PointNeXt, we use the PointNeXt-XL architecture pre-trained on S3DIS \cite{s3dis} with a segmentation objective. We report the results on area1 as it performs best. For further details see App.~\ref{app:method_impl}.

\begin{figure}[t]

\begin{subfigure}{0.5\textwidth}
\includegraphics[width=0.9\linewidth]{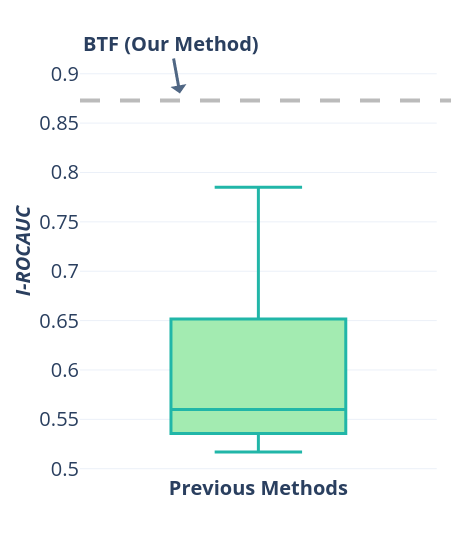}
\caption{\textit{\textbf{Summary I-ROCAUC Results}}}
\label{fig:image_rocauc}
\end{subfigure}
\begin{subfigure}{0.5\textwidth}
\includegraphics[width=0.9\linewidth]{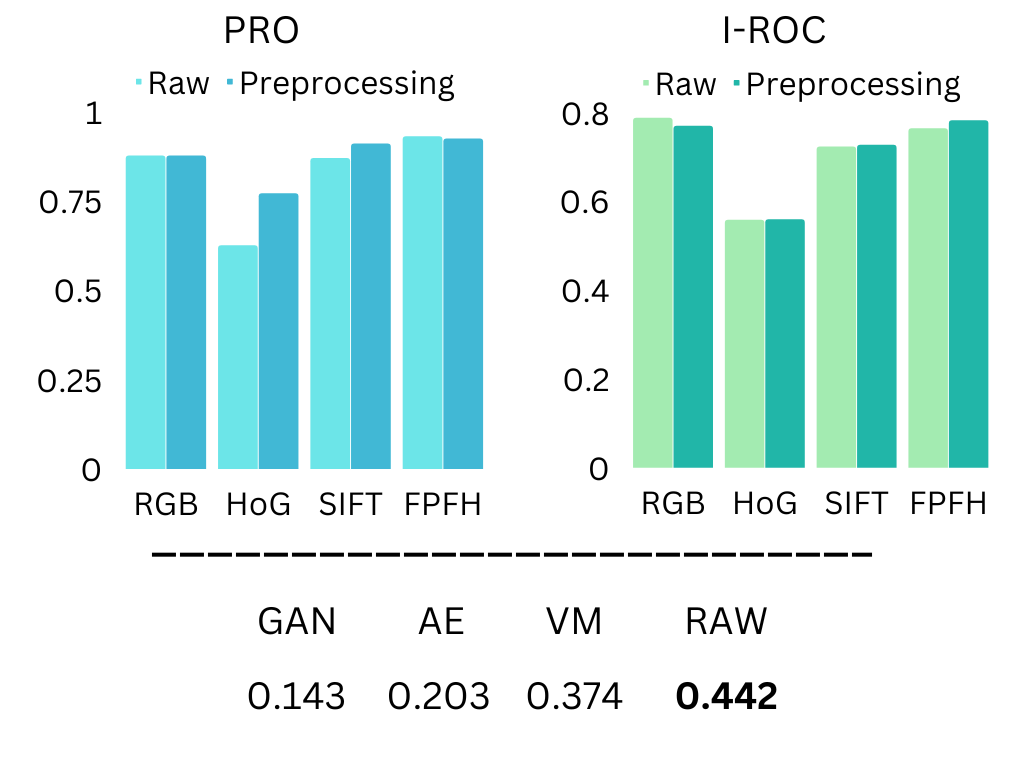}
\caption{\textit{\textbf{Preprocessing}}}
\label{fig:preprocessing_ablation}
\end{subfigure}
\caption{\textit{\textbf{Summary I-ROCAUC Results (left):}} Our proposed BTF (top, dashed line) outperforms all $21$ other methods (bottom, candle chart). For the numbers, see App.~\ref{app:iroc_results}.\\ \textit{\textbf{Preprocessing (right):}} Top: comparing the raw and preprocessed data. Depth-based methods (i.e. HoG and SIFT) benefit from the preprocessing on anomaly segmentation (\textit{PRO}), point-cloud based methods (i.e. FPFH) benefit from the preprocessing on anomaly detection (\textit{I-ROC}).  Bottom: \textit{PRO} results of MVTec 3D-AD baselines vs. preprocessed RAW representation, the preprocessed RAW outperforms the MVTec 3D-AD baselines. Average metrics across all classes are reported
}
\end{figure}

\noindent \textbf{Establishing a 3D-based preprocessing protocol.} Preprocessing is sometimes required for removing nuisance artifacts. To handle such cases we developed a simple preprocessing method. We first remove the background plane by applying RANSAC \cite{ransac} on the point cloud data. Once removed, we discard outliers and areas far from the plane by applying a connected-components-based algorithm (for implementation details see App.~\ref{app:preprocessing_impl}).
This preprocessing phase left the results of color-only methods mostly unaffected. More interestingly, it drastically improved results for the depth-based methods, while for 3D-based methods (i.e. FPFH) it slightly decreases results. We postulate this is caused by the difference in how depth and 3D methods handle missing sensor information\footnote{3D sensing methods are prone to sampling noise and missing information (e.g. occlusions). In MVTec 3D-AD it is common to have very noisy backgrounds, these areas are replaced by zeros by the dataset designers.}. For point clouds, these missing values are all located at the origin (since their value is $0$) and are easily ignored (since they are not in the spatial context of other points). In contrast, for depth images, these values are in the spatial context of other points and are thus taken into account. Removing these planes creates a similar situation to that of the point clouds and hence benefits depth-based methods. When using the preprocessed data, even the simplest feature (i.e. Raw) outperforms the original baselines \cite{mvtec3d}, results are shown in Fig.~\ref{fig:preprocessing_ablation}.

\subsection{Limitations.} \noindent Our proposed method BTF has several limitations:

\noindent \textbf{Feature fusion.} Both \textit{cable gland} and \textit{foam} perform poorly for all depth-based methods (Tab.~\ref{table:au_pro} and App.~\ref{app:iroc_results}, App.~\ref{app:proc_results}). While the anomalies in these classes are easier to detect by using color than by using 3D (see Fig.~\ref{fig:rgb_v_d}), we expected the fusion of both modalities to improve performance. Unfortunately, for these classes, the fused features underperformed the color-only method. Future work should address this issue. 

\noindent  \textbf{Image-level accuracy.} While \textit{BTF} establishes a new state-of-the-art on all metrics, the image level detection accuracy is far from perfect. It reaches an \textit{I-ROC} of $86.5\%$, a large improvement compared to past methods, but still a relatively low score. Since we use PatchCore as the backbone for most of our experiments, the \textit{I-ROC} score is determined by the image patch that is most distant from all training patches. We expect that better metrics can be devised for 3D data; investigating them is left for future work.



\section{Conclusion}
Our study was motivated by the outperformance of color-only approaches over all existing 3D methods on the MVTec 3D-AD dataset. We conducted an extensive investigation of 3D representations and found that rotation-invariant representations achieves the best performance on 3D anomaly detection. We proposed \textit{BTF}, a combination of 3D and color features that set a new state-of-the-art. As our method is simple, we expect it to serve as a strong baseline for future work.

\section{Acknowledgements}
This work was supported in part by Oracle Cloud credits and related resources provided by the Oracle for Research program.

\clearpage
\bibliographystyle{splncs04}
\bibliography{egbib}

\begin{thebibliography}{10}
\providecommand{\url}[1]{\texttt{#1}}
\providecommand{\urlprefix}{URL }
\providecommand{\doi}[1]{https://doi.org/#1}

\bibitem{spinnet}
Ao, S., Hu, Q., Yang, B., Markham, A., Guo, Y.: Spinnet: Learning a general
  surface descriptor for 3d point cloud registration. In: Proceedings of the
  IEEE/CVF Conference on Computer Vision and Pattern Recognition (2021)

\bibitem{s3dis}
Armeni, I., Sener, O., Zamir, A.R., Jiang, H., Brilakis, I., Fischer, M.,
  Savarese, S.: 3d semantic parsing of large-scale indoor spaces. In:
  Proceedings of the IEEE Conference on Computer Vision and Pattern Recognition
  (CVPR) (June 2016)

\bibitem{bengs2021three}
Bengs, M., Behrendt, F., Kr{\"u}ger, J., Opfer, R., Schlaefer, A.:
  Three-dimensional deep learning with spatial erasing for unsupervised anomaly
  segmentation in brain mri. International journal of computer assisted
  radiology and surgery  \textbf{16}(9),  1413--1423 (2021)

\bibitem{mvtec2d}
Bergmann, P., Fauser, M., Sattlegger, D., Steger, C.: Mvtec ad--a comprehensive
  real-world dataset for unsupervised anomaly detection. In: Proceedings of the
  IEEE/CVF conference on computer vision and pattern recognition. pp.
  9592--9600 (2019)

\bibitem{bergmann2020uninformed}
Bergmann, P., Fauser, M., Sattlegger, D., Steger, C.: Uninformed students:
  Student-teacher anomaly detection with discriminative latent embeddings. In:
  Proceedings of the IEEE/CVF Conference on Computer Vision and Pattern
  Recognition. pp. 4183--4192 (2020)

\bibitem{mvtec3d}
Bergmann, P., Jin, X., Sattlegger, D., Steger, C.: The mvtec 3d-ad dataset for
  unsupervised 3d anomaly detection and localization. arXiv preprint
  arXiv:2112.09045  (2021)

\bibitem{bergmann2022anomaly}
Bergmann, P., Sattlegger, D.: Anomaly detection in 3d point clouds using deep
  geometric descriptors. arXiv preprint arXiv:2202.11660  (2022)

\bibitem{3d_ad_pc_deep}
Bergmann, P., Sattlegger, D.: Anomaly detection in 3d point clouds using deep
  geometric descriptors. arXiv preprint arXiv:2202.11660  (2022)

\bibitem{rocauc}
Bradley, A.P.: The use of the area under the roc curve in the evaluation of
  machine learning algorithms. Pattern recognition  \textbf{30}(7),  1145--1159
  (1997)

\bibitem{chen2020simple}
Chen, T., Kornblith, S., Norouzi, M., Hinton, G.: A simple framework for
  contrastive learning of visual representations. arXiv preprint
  arXiv:2002.05709  (2020)

\bibitem{spade}
Cohen, N., Hoshen, Y.: Sub-image anomaly detection with deep pyramid
  correspondences. arXiv preprint arXiv:2005.02357  (2020)

\bibitem{hog}
Dalal, N., Triggs, B.: Histograms of oriented gradients for human detection.
  In: 2005 IEEE computer society conference on computer vision and pattern
  recognition (CVPR'05). vol.~1, pp. 886--893. Ieee (2005)

\bibitem{defard2021padim}
Defard, T., Setkov, A., Loesch, A., Audigier, R.: Padim: a patch distribution
  modeling framework for anomaly detection and localization. In: International
  Conference on Pattern Recognition. pp. 475--489. Springer (2021)

\bibitem{imagenet}
Deng, J., Dong, W., Socher, R., Li, L.J., Li, K., Fei-Fei, L.: Imagenet: A
  large-scale hierarchical image database. In: 2009 IEEE conference on computer
  vision and pattern recognition. pp. 248--255. Ieee (2009)

\bibitem{eskin2002geometric}
Eskin, E., Arnold, A., Prerau, M., Portnoy, L., Stolfo, S.: A geometric
  framework for unsupervised anomaly detection. In: Applications of data mining
  in computer security, pp. 77--101. Springer (2002)

\bibitem{dbscan}
Ester, M., Kriegel, H.P., Sander, J., Xu, X., et~al.: A density-based algorithm
  for discovering clusters in large spatial databases with noise. In: KDD
  (1996)

\bibitem{ransac}
Fischler, M.A., Bolles, R.C.: Random sample consensus: a paradigm for model
  fitting with applications to image analysis and automated cartography.
  Communications of the ACM  \textbf{24}(6),  381--395 (1981)

\bibitem{gidaris2018unsupervised}
Gidaris, S., Singh, P., Komodakis, N.: Unsupervised representation learning by
  predicting image rotations. arXiv preprint arXiv:1803.07728  (2018)

\bibitem{glodek2013ensemble}
Glodek, M., Schels, M., Schwenker, F.: Ensemble gaussian mixture models for
  probability density estimation. Computational Statistics  \textbf{28}(1),
  127--138 (2013)

\bibitem{golan2018deep}
Golan, I., El-Yaniv, R.: Deep anomaly detection using geometric
  transformations. In: NeurIPS (2018)

\bibitem{he2019moco}
He, K., Fan, H., Wu, Y., Xie, S., Girshick, R.: Momentum contrast for
  unsupervised visual representation learning. arXiv preprint arXiv:1911.05722
  (2019)

\bibitem{he2020momentum}
He, K., Fan, H., Wu, Y., Xie, S., Girshick, R.: Momentum contrast for
  unsupervised visual representation learning. In: Proceedings of the IEEE/CVF
  Conference on Computer Vision and Pattern Recognition. pp. 9729--9738 (2020)

\bibitem{hendrycks2019using}
Hendrycks, D., Mazeika, M., Kadavath, S., Song, D.: Using self-supervised
  learning can improve model robustness and uncertainty. In: NeurIPS (2019)

\bibitem{jolliffe2011principal}
Jolliffe, I.: Principal component analysis. Springer (2011)

\bibitem{latecki2007outlier}
Latecki, L.J., Lazarevic, A., Pokrajac, D.: Outlier detection with kernel
  density functions. In: International Workshop on Machine Learning and Data
  Mining in Pattern Recognition. pp. 61--75. Springer (2007)

\bibitem{cutpaste}
Li, C.L., Sohn, K., Yoon, J., Pfister, T.: Cutpaste: Self-supervised learning
  for anomaly detection and localization. In: Proceedings of the IEEE/CVF
  Conference on Computer Vision and Pattern Recognition. pp. 9664--9674 (2021)

\bibitem{liu2008isolation}
Liu, F.T., Ting, K.M., Zhou, Z.H.: Isolation forest. In: 2008 Eighth IEEE
  International Conference on Data Mining. pp. 413--422. IEEE (2008)

\bibitem{sift}
Lowe, D.G.: Distinctive image features from scale-invariant keypoints.
  International journal of computer vision  \textbf{60}(2),  91--110 (2004)

\bibitem{perera2019learning}
Perera, P., Patel, V.M.: Learning deep features for one-class classification.
  IEEE Transactions on Image Processing  \textbf{28}(11),  5450--5463 (2019)

\bibitem{poisson_blending}
P{\'e}rez, P., Gangnet, M., Blake, A.: Poisson image editing. SIGGRAPH  (2003)

\bibitem{point_next}
Qian, G., Li, Y., Peng, H., Mai, J., Hammoud, H., Elhoseiny, M., Ghanem, B.:
  Pointnext: Revisiting pointnet++ with improved training and scaling
  strategies. arXiv:2206.04670  (2022)

\bibitem{panda}
Reiss, T., Cohen, N., Bergman, L., Hoshen, Y.: Panda: Adapting pretrained
  features for anomaly detection and segmentation. In: Proceedings of the
  IEEE/CVF Conference on Computer Vision and Pattern Recognition. pp.
  2806--2814 (2021)

\bibitem{unet}
Ronneberger, O., Fischer, P., Brox, T.: U-net: Convolutional networks for
  biomedical image segmentation. In: International Conference on Medical image
  computing and computer-assisted intervention. pp. 234--241. Springer (2015)

\bibitem{patchcore}
Roth, K., Pemula, L., Zepeda, J., Sch{\"o}lkopf, B., Brox, T., Gehler, P.:
  Towards total recall in industrial anomaly detection. arXiv preprint
  arXiv:2106.08265  (2021)

\bibitem{ruff2018deep}
Ruff, L., Gornitz, N., Deecke, L., Siddiqui, S.A., Vandermeulen, R., Binder,
  A., M{\"u}ller, E., Kloft, M.: Deep one-class classification. In: ICML (2018)

\bibitem{fpfh}
Rusu, R.B., Blodow, N., Beetz, M.: Fast point feature histograms (fpfh) for 3d
  registration. In: 2009 IEEE International Conference on Robotics and
  Automation. pp. 3212--3217 (2009). \doi{10.1109/ROBOT.2009.5152473}

\bibitem{schlegl2019f}
Schlegl, T., Seeb{\"o}ck, P., Waldstein, S.M., Langs, G., Schmidt-Erfurth, U.:
  f-anogan: Fast unsupervised anomaly detection with generative adversarial
  networks. Medical image analysis  \textbf{54},  30--44 (2019)

\bibitem{schlegl2017unsupervised}
Schlegl, T., Seeb{\"o}ck, P., Waldstein, S.M., Schmidt-Erfurth, U., Langs, G.:
  Unsupervised anomaly detection with generative adversarial networks to guide
  marker discovery. In: International Conference on Information Processing in
  Medical Imaging (2017)

\bibitem{nsa}
Schl{\"u}ter, H.M., Tan, J., Hou, B., Kainz, B.: Self-supervised
  out-of-distribution detection and localization with natural synthetic
  anomalies (nsa). arXiv preprint arXiv:2109.15222  (2021)

\bibitem{scholkopf2000support}
Scholkopf, B., Williamson, R.C., Smola, A.J., Shawe-Taylor, J., Platt, J.C.:
  Support vector method for novelty detection. In: NIPS (2000)

\bibitem{simarro2020unsupervised}
Simarro~Viana, J., de~la Rosa, E., Vande~Vyvere, T., Robben, D., Sima, D.M.,
  et~al.: Unsupervised 3d brain anomaly detection. In: International MICCAI
  Brainlesion Workshop. pp. 133--142. Springer (2020)

\bibitem{tack2020csi}
Tack, J., Mo, S., Jeong, J., Shin, J.: Csi: Novelty detection via contrastive
  learning on distributionally shifted instances. arXiv preprint
  arXiv:2007.08176  (2020)

\bibitem{modelnet10}
Wu, Z., Song, S., Khosla, A., Yu, F., Zhang, L., Tang, X., Xiao, J.: 3d
  shapenets: A deep representation for volumetric shapes. In: Proceedings of
  the IEEE conference on computer vision and pattern recognition. pp.
  1912--1920 (2015)

\bibitem{yu2021fastflow}
Yu, J., Zheng, Y., Wang, X., Li, W., Wu, Y., Zhao, R., Wu, L.: Fastflow:
  Unsupervised anomaly detection and localization via 2d normalizing flows.
  arXiv preprint arXiv:2111.07677  (2021)

\bibitem{wide_resnet_50}
Zagoruyko, S., Komodakis, N.: Wide residual networks. arXiv preprint
  arXiv:1605.07146  (2016)

\bibitem{3dmatch}
Zeng, A., Song, S., Nie{\ss}ner, M., Fisher, M., Xiao, J., Funkhouser, T.:
  3dmatch: Learning local geometric descriptors from rgb-d reconstructions. In:
  CVPR (2017)

\bibitem{open3d}
Zhou, Q.Y., Park, J., Koltun, V.: {Open3D}: {A} modern library for {3D} data
  processing. arXiv:1801.09847  (2018)

\bibitem{zong2018deep}
Zong, B., Song, Q., Min, M.R., Cheng, W., Lumezanu, C., Cho, D., Chen, H.: Deep
  autoencoding gaussian mixture model for unsupervised anomaly detection. ICLR
  (2018)

\end{thebibliography}

\appendix

\clearpage

\section{Detailed \textit{I-ROC} Results}
\label{app:iroc_results}
\begin{table}[h!]
\begin{center}
\caption{\textbf{\textit{Detailed I-ROCAUC Results:}} Top half are current state-of-the-art, bottom half are methods investigated by us. A large number of our methods outperform all current methods by a wide margin. ``iNet`` indicates ImageNet pre-trained}
\label{table:image_rocauc}
\resizebox{\linewidth}{!}{%
\begin{tabular}{@{\hskip5pt}l@{\hskip5pt}l|cccccccccc|c} 
    \toprule
    & Method & Bagel & \begin{tabular}[c]{@{}c@{}}Cable\\ Gland\end{tabular} & Carrot & Cookie & Dowel & Foam & Peach & Potato & Rope & Tire & Mean \\ 
    \midrule
    \multirow{12}{*}{\rotatebox[origin=c]{90}{Previous Methods}} & Voxel GAN & 0.383 & 0.623  & 0.474 & 0.639 & 0.564 & 0.409 & 0.617 & 0.427 & 0.663 & 0.577 & 0.537 \\
    & \qquad \qquad  + RGB & 0.680 & 0.324  & 0.565 & 0.399 & 0.497 & 0.482 & 0.566 & 0.579 & 0.601 & 0.482 & 0.517 \\
    & Voxel AE & 0.693 & 0.425  & 0.515 & 0.79 & 0.494 & 0.558 & 0.537 & 0.484 & 0.639 & 0.583 & 0.571 \\
    & \qquad \qquad  + RGB & 0.510 & 0.540  & 0.384 & 0.693 & 0.446 & 0.632 & 0.550 & 0.494 & 0.721 & 0.413 & 0.538 \\
    & Voxel VM & 0.75 & 0.747  & 0.613 & 0.738 & 0.823 & 0.693 & 0.679 & 0.652 & 0.609 & 0.69 & 0.699 \\ 
    & \qquad \qquad  + RGB & 0.553 & 0.772  & 0.484 & 0.701 & 0.751 & 0.578 & 0.480 & 0.466 & 0.689 & 0.611 & 0.609 \\
    \cmidrule{2-13}
    & Depth GAN & 0.53 & 0.376  & 0.607 & 0.603 & 0.497 & 0.484 & 0.595 & 0.489 & 0.536 & 0.521 & 0.523 \\
    & \qquad \qquad  + RGB & 0.538 & 0.372  & 0.580 & 0.603 & 0.430 & 0.534 & 0.642 & 0.601 & 0.443 & 0.577 & 0.532 \\
    & Depth AE & 0.468 & 0.731  & 0.497 & 0.673 & 0.534 & 0.417 & 0.485 & 0.549 & 0.564 & 0.546 & 0.546 \\
    & \qquad \qquad  + RGB & 0.648 & 0.502  & 0.650 & 0.488 & 0.805 & 0.522 & 0.712 & 0.529 & 0.540 & 0.552 & 0.595 \\
    & Depth VM &  0.51 & 0.542  & 0.469 & 0.576 & 0.609 & 0.699 & 0.45 & 0.419 & 0.668 & 0.52 & 0.546 \\
    & \qquad \qquad  + RGB & 0.513 & 0.551  & 0.477 & 0.581 & 0.617 & 0.716 & 0.450 & 0.421 & 0.598 & 0.623 & 0.555 \\
    \midrule
    \multirow{10}{*}{\rotatebox[origin=c]{90}{Our Findings}} & RGB iNet & 0.854 & \textbf{0.840} & 0.824 & 0.687 & \textbf{0.974} & 0.716 &  0.713 & 0.593 & 0.920 & 0.724 & 0.785 \\
    & Depth iNet & 0.624 & 0.683 & 0.676 & 0.838 & 0.608 & 0.558 &  0.567 & 0.496 & 0.699 & 0.619 & 0.637 \\
    & NSA & 0.841  &  0.494 &  0.776 & 0.913  & 0.636 & 0.616 & 0.795 & 0.597 & 0.856 & 0.438 & 0.696 \\
    & Raw & 0.578 & 0.732 & 0.444 & 0.798 & 0.579 & 0.537 &  0.347 & 0.306 & 0.439 & 0.517 & 0.528 \\
    & HoG & 0.560 & 0.615 & 0.676 & 0.491 & 0.598 & 0.489 &  0.542 & 0.553 & 0.655 & 0.423 & 0.560 \\ 
    & SIFT & 0.696 & 0.553 & 0.824 & 0.696 & 0.795 & \textbf{0.773} &  0.573 & 0.746 & 0.936 & 0.553 & 0.714 \\
    & FPFH & 0.820 & 0.533 & 0.877 & 0.769 & 0.718 & 0.574 &  0.774 & 0.895 & \textbf{0.990} & 0.582 & 0.753 \\
    & PointNext (area 1) & 0.499 & 0.772 & 0.498 & 0.750 & 0.589 & 0.525 & 0.545 & 0.431 & 0.805 & 0.484 & 0.587 \\
    & SpinNet & 0.535 & 0.413 & 0.568 & 0.662 & 0.472 & 0.480 & 0.367 & 0.494 & 0.722 & 0.527 & 0.524 \\
    & \textit{BTF} & \textbf{0.938} & 0.765 & \textbf{0.972} & \textbf{0.888} & 0.960 & 0.664 &  \textbf{0.904} & \textbf{0.929} & 0.982 & \textbf{0.726} & \textbf{0.873} \\
  \bottomrule
    \end{tabular}
}
\end{center}
\end{table}

\newpage

\section{Detailed \textit{P-ROC} Results}
\label{app:proc_results}
\begin{table}[h!]
\begin{center}
\caption{\textbf{\textit{Detailed P-ROC Results:}} Results not reported for previous methods, "iNet" indicates ImageNet pre-trained}
\label{table:pixel_rocauc}
\resizebox{\linewidth}{!}{%
\begin{tabular}{l|cccccccccc|c} 
    \toprule
     Method & Bagel & \begin{tabular}[c]{@{}c@{}}Cable\\ Gland\end{tabular} & Carrot & Cookie & Dowel & Foam & Peach & Potato & Rope & Tire & Mean \\ 
    \midrule
    RGB iNet & 0.983 & 0.984 & 0.980 & 0.974 & 0.985 & 0.836 & 0.976 & 0.982 & 0.989 & 0.975 & 0.966 \\
    Depth iNet & 0.941 & 0.759 & 0.933 & 0.946 & 0.829 & 0.518 & 0.939 & 0.743 & 0.974 & 0.632 & 0.821 \\
    NSA & 0.925  &  0.638 &  0.872 & 0.908  & 0.674 & 0.777 & 0.902 & 0.825 & 0.972 & 0.676 & 0.817 \\
    Raw Depth & 0.404 & 0.306 & 0.772 & 0.457 & 0.641 & 0.478 & 0.354 & 0.602 & 0.905 & 0.558 & 0.548 \\
    HoG & 0.782 & 0.846 & 0.965 & 0.684 & 0.848 & 0.741 & 0.779 & 0.973 & 0.926 & 0.903 & 0.845 \\ 
    SIFT & 0.974 & 0.862 & 0.993 & 0.952 & 0.980 & 0.862 & 0.955 & 0.996 & 0.993 & 0.971 & 0.954 \\
    FPFH &  0.995 & 0.955 & \textbf{0.998} & 0.971 & 0.993 & 0.911 & 0.995 & \textbf{0.999} & \textbf{0.998} & 0.988 & 0.980 \\
    Omnivore & 0.936 & 0.840 & 0.776 & 0.901 & 0.919 & 0.850 & 0.894 & 0.911 & 0.981 & 0.958 & 0.896 \\
    PointNext & 0.735 & 0.652 & 0.708 & 0.899 & 0.640 & 0.481 & 0.769 & 0.384 & 0.959 & 0.651 & 0.687 \\
    SpinNet & 0.882 & 0.684 & 0.978 & 0.902 & 0.963 & 0.771 & 0.833 & 0.911 & 0.994 & 0.817 & 0.873 \\
     \textit{BTF} & \textbf{0.996} & \textbf{0.991} & 0.997 & \textbf{0.995} & \textbf{0.995} & \textbf{0.972} & \textbf{0.996} & 0.998 & 0.995 & \textbf{0.994} & \textbf{0.993} \\
  \bottomrule
    \end{tabular}
}
\end{center}
\end{table}

\section{Detailed PointNext Results}
\label{app:point_next_results}
Tab.~\ref{table:au_pro_point_next}, \ref{table:image_rocauc_point_next}, and \ref{table:pixel_rocauc_point_next} contain the full breakdown of the PointNext \cite{point_next} results on the S3DIS \cite{s3dis} dataset. All the results are based on the PointNext-XL model. PointNext trained a different model for each of the six areas in S3DIS.

\begin{table}[h!]
\begin{center}
\caption{\textbf{\textit{Detailed PointNext PRO Results}}}
\label{table:au_pro_point_next}
\resizebox{\linewidth}{!}{%
\begin{tabular}{c|cccccccccc|c} 
    \toprule
     \begin{tabular}[c]{@{}c@{}}Pretrained\\ Area\end{tabular} & Bagel & \begin{tabular}[c]{@{}c@{}}Cable\\ Gland\end{tabular} & Carrot & Cookie & Dowel & Foam & Peach & Potato & Rope & Tire & Mean \\ 
    \midrule
    Area1 & 0.425 & 0.294 & 0.365 & \textbf{0.772} & \textbf{0.227} & 0.151 & \textbf{0.408} & 0.101 & 0.771 & 0.295 & \textbf{0.380} \\
    Area2 & 0.392 & 0.261 & 0.282 & 0.549 & 0.183 & 0.166 & 0.286 & 0.168 & 0.788 & 0.293 & 0.336 \\
    Area3 & \textbf{0.484} & 0.299 & 0.295 & 0.671 & 0.205 & 0.161 & 0.387 & \textbf{0.173} & 0.690 & 0.313 & 0.367 \\
    Area4 & 0.302 & 0.309 & 0.314 & 0.554 & 0.185 & 0.158 & 0.253 & 0.164 & 0.726 & \textbf{0.343} & 0.330 \\
    Area5 & 0.282 & \textbf{0.315} & 0.268 & 0.649 & 0.202 & \textbf{0.173} & 0.263 & 0.116 & 0.522 & 0.312 & 0.310 \\
    Area6 & 0.355 & 0.305 & \textbf{0.410} & 0.633 & 0.189 & 0.147 & 0.305 & 0.157 & \textbf{0.799} & 0.262 & 0.356 \\
    
  \bottomrule
    \end{tabular}
}
\end{center}
\end{table}

\begin{table}[h!]
\begin{center}
\caption{\textbf{\textit{Detailed PointNext I-ROC Results}}}
\label{table:image_rocauc_point_next}
\resizebox{\linewidth}{!}{%
\begin{tabular}{c|cccccccccc|c} 
    \toprule
     \begin{tabular}[c]{@{}c@{}}Pretrained\\ Area\end{tabular} & Bagel & \begin{tabular}[c]{@{}c@{}}Cable\\ Gland\end{tabular} & Carrot & Cookie & Dowel & Foam & Peach & Potato & Rope & Tire & Mean \\ 
    \midrule
    Area1 & 0.499 & \textbf{0.772} & 0.498 & 0.750 & \textbf{0.589} & \textbf{0.525} & \textbf{0.545} & \textbf{0.431} & 0.805 & 0.484 & \textbf{0.587} \\
    Area2 & 0.565 & 0.600 & 0.516 & 0.486 & 0.541 & 0.379 & 0.476 & 0.308 & 0.760 & 0.432 & 0.506 \\
    Area3 & \textbf{0.614} & 0.697 & 0.489 & 0.588 & 0.571 & 0.446 & 0.476 & 0.318 & 0.763 & 0\textbf{.516} & 0.547 \\
    Area4 & 0.536 & 0.698 & 0.496 & 0.491 & 0.538 & 0.459 & 0.487 & 0.275 & 0.776 & 0.501 & 0.525 \\
    Area5 & 0.585 & 0.641 & \textbf{0.563} & \textbf{0.831} & 0.555 & 0.412 & 0.413 & 0.342 & \textbf{0.816} & 0.459 & 0.561 \\
    Area6 & 0.536 & 0.674 & 0.547 & 0.620 & 0.517 & 0.406 & 0.497 & 0.319 & 0.793 & 0.493 & 0.540 \\
    
  \bottomrule
    \end{tabular}
}
\end{center}
\end{table}

\begin{table}[h!]
\begin{center}
\caption{\textbf{\textit{Detailed PointNext P-ROC Results}}}
\label{table:pixel_rocauc_point_next}
\resizebox{\linewidth}{!}{%
\begin{tabular}{c|cccccccccc|c} 
    \toprule
     \begin{tabular}[c]{@{}c@{}}Pretrained\\ Area\end{tabular}  & Bagel & \begin{tabular}[c]{@{}c@{}}Cable\\ Gland\end{tabular} & Carrot & Cookie & Dowel & Foam & Peach & Potato & Rope & Tire & Mean \\ 
    \midrule
    Area1 & 0.735 & 0.652 & 0.708 & \textbf{0.899} & \textbf{0.640} & 0.481 & \textbf{0.769}& 0.384 & 0.959 & 0.651 & \textbf{0.687} \\
    Area2 & 0.689 & 0.650 & 0.644 & 0.780 & 0.515 & 0.425 & 0.656 & \textbf{0.478} & \textbf{0.964} & 0.643 & 0.644 \\
    Area3 & \textbf{0.762} & 0.660 & 0.688 & 0.845 & 0.475 & 0.505 & 0.755 & 0.469 & 0.945 & 0.654 & 0.675 \\
    Area4 & 0.648 & 0.643 & 0.678 & 0.807 & 0.545 & \textbf{0.512} & 0.628 & 0.437 & 0.951 & \textbf{0.715} & 0.656 \\
    Area5 & 0.666 & \textbf{0.681} & 0.645 & 0.837 & 0.574 & 0.507 & 0.668 & 0.385 & 0.898 & 0.661 & 0.652 \\
    Area6 & 0.716 & 0.577 & \textbf{0.726} & 0.837 & 0.589 & \textbf{0.512} & 0.672 & 0.425 & 0.962 & 0.632 & 0.664 \\
    
  \bottomrule
    \end{tabular}
}
\end{center}
\end{table}

\section{Additional Method Combinations}
\label{app:rgb_depth_comb}
Tab.~\ref{table:au_pro_with_rgb}, \ref{table:image_rocauc_with_rgb}, and \ref{table:pixel_rocauc_with_rgb} contain the results when combining depth and RGB from other methods, not shown in the main paper.

\begin{table}[h!]
\begin{center}
\caption{\textbf{\textit{Additional Method Combinations PRO Results}}}
\label{table:au_pro_with_rgb}
\resizebox{\linewidth}{!}{%
\begin{tabular}{l|cccccccccc|c} 
    \toprule
     Method & Bagel & \begin{tabular}[c]{@{}c@{}}Cable\\ Gland\end{tabular} & Carrot & Cookie & Dowel & Foam & Peach & Potato & Rope & Tire & Mean \\ 
    \midrule
    Depth iNet+RGB & 0.877 & 0.893 & 0.908 & 0.924 & 0.877 & 0.464 & 0.927 & 0.929 & 0.911 & 0.829 & 0.853 \\
    Raw+RGB & 0.896 & 0.948 & 0.927 & 0.874 & 0.925 & 0.549 & 0.903 & 0.932 & 0.910 & 0.887 & 0.875 \\
    HoG+RGB & 0.898 & 0.948 & 0.927 & 0.873 & 0.927 & 0.555 & 0.902 & 0.932 & 0.911 & 0.901 & 0.877 \\
    SIFT+RGB & 0.895 & 0.947 & 0.927 & 0.875 & 0.929 & 0.555 & 0.904 & 0.932 & 0.910 & 0.895 & 0.876 \\
    SpinNet+RGB & 0.897 & 0.948 & 0.929 & 0.878 & 0.928 & 0.550 & 0.904 & 0.931 & 0.911 & 0.899 & 0.877 \\
    
  \bottomrule
    \end{tabular}
}
\end{center}
\end{table}

\begin{table}[h!]
\begin{center}
\caption{\textbf{\textit{Additional Method Combinations I-ROC Results}}}
\label{table:image_rocauc_with_rgb}
\resizebox{\linewidth}{!}{%
\begin{tabular}{l|cccccccccc|c} 
    \toprule
     Method & Bagel & \begin{tabular}[c]{@{}c@{}}Cable\\ Gland\end{tabular} & Carrot & Cookie & Dowel & Foam & Peach & Potato & Rope & Tire & Mean \\ 
    \midrule
    Depth iNet+RGB & 0.808 & 0.707 & 0.739 & 0.836 & 0.882 & 0.547 & 0.731 & 0.667 & 0.825 & 0.648 & 0.739 \\
    Raw+RGB & 0.877 & 0.876 & 0.785 & 0.718 & 0.960 & 0.699 & 0.742 & 0.581 & 0.895 & 0.623 & 0.775 \\
    HoG+RGB & 0.887 & 0.891 & 0.791 & 0.716 & 0.972 & 0.676 & 0.714 & 0.576 & 0.862 & 0.649 & 0.773 \\
    SIFT+RGB & 0.845 & 0.882 & 0.780 & 0.727 & 0.966 & 0.671 & 0.726 & 0.619 & 0.867 & 0.681 & 0.776 \\
    SpinNet+RGB & 0.851 & 0.841 & 0.806 & 0.682 & 0.969 & 0.753 & 0.713 & 0.627 & 0.864 & 0.679 & 0.778 \\
    
  \bottomrule
    \end{tabular}
}
\end{center}
\end{table}

\begin{table}[h!]
\begin{center}
\caption{\textbf{\textit{Additional Method Combinations P-ROC Results}}}
\label{table:pixel_rocauc_with_rgb}
\resizebox{\linewidth}{!}{%
\begin{tabular}{l|cccccccccc|c} 
    \toprule
     Method & Bagel & \begin{tabular}[c]{@{}c@{}}Cable\\ Gland\end{tabular} & Carrot & Cookie & Dowel & Foam & Peach & Potato & Rope & Tire & Mean \\ 
    \midrule
    Depth iNet+RGB & 0.981 & 0.966 & 0.973 & 0.983 & 0.971 & 0.777 & 0.983 & 0.979 & 0.988 & 0.954 & 0.955 \\
    Raw+RGB & 0.983 & 0.984 & 0.980 & 0.974 & 0.985 & 0.829 & 0.976 & 0.982 & 0.987 & 0.972 & 0.965 \\
    HoG+RGB & 0.983 & 0.984 & 0.980 & 0.974 & 0.985 & 0.832 & 0.976 & 0.982 & 0.987 & 0.976 & 0.965 \\
    SIFT+RGB & 0.983 & 0.984 & 0.980 & 0.974 & 0.985 & 0.829 & 0.976 & 0.982 & 0.987 & 0.975 & 0.965 \\
    SpinNet+RGB & 0.983 & 0.984 & 0.980 & 0.975 & 0.985 & 0.830 & 0.976 & 0.982 & 0.987 & 0.976 & 0.965 \\
    
  \bottomrule
    \end{tabular}
}
\end{center}
\end{table}

\section{Method Specific Implementation Details}
\label{app:method_impl}
Bellow, we present additional implementation details for each of our investigated methods

\subsection{RGB-only ImageNet Features} Utilizing PatchCore, we feed RGB images an ImageNet \cite{imagenet} pre-trained WideResNet50 \cite{wide_resnet_50} backbone as a feature extractor. To allow for localized segmentation, we extract patch-level features from the aggregated outputs of blocks $2$ and $3$, resulting in a feature dimension of $1536$. 

\subsection{Depth-only ImageNet Features} As in the RGB-only case, we use PatchCore with a depth map normalized according to ImageNet statistics.

\subsection{Raw Depth Values} We divide the depth image into patches of $8\times{8}$ pixels, resulting in $28\times{28}$ patches. The descriptor is comprised of the $8\times{8}$ pixels of each patch, flattened to a 1D list of length $64$.

\subsection{NSA} We were not able to find an official implementation of CutPaste \cite{cutpaste} and the public unofficial implementations lag behind the reported figures by up to $10\%$. Therefore, we compare our method to NSA \cite{nsa}, a follow-up to CutPaste that uses Poisson blending \cite{poisson_blending} to achieve more realistic augmentations. To test NSA on the new dataset, we modified their official implementation to handle depth images. The current implementation requires the images to be represented as integers, as such, the depth images are discretized to [0, 255]. Furthermore, NSA uses extensive, class-dependent hyperparameters. MVTec 3D-AD classes were assigned hyperparameters by visually comparing them with the original classes and assigning the values of the most similar class. We use depth images for running these experiments. It is possible that NSA performance can be improved by specific per-class augmentations, however, this requires prior knowledge of anomalies.

\subsection{HoG} We use the depth image as input. To align with the feature map resolution, we use $8$ pixels per cell, and $1$ cell per block. We use $8$ bins to arrive at a $32$ dimensional representation.

\subsection{D-SIFT} We use the depth image as input. We apply Dense SIFT to all the pixels, to reduce the resolution, we apply average pooling. Following standard SIFT practice, we use a $128$ feature dimension.

\subsection{FPFH}
To speed up calculations of FPFH, we downsampled the point cloud prior to running the algorithm. The downsampling is performed on the organized point cloud (i.e. image downsampling); it is then flattened into an unorganized point cloud. Using the implementation from the open-source library Open3D \cite{open3d}, we extract a descriptor for each point. We then reshape these descriptors back into an organized point cloud and their resolution is lowered by average pooling them. FPFH requires normals to run, we estimate the normals using Open3D. The radius for the FPFH algorithm is $0.25$ and the $max\_nn$ parameter is set to $100$. The resulting feature is of dimension $33$.

\subsection{PointNext}
We use tgethe PointNeXt-XL variant to extract features. To overcome the limitation of using a very small number of points ($1024$ or $2048$), we use the S3DIS \cite{s3dis} pretrained model. Meaning, the model was pretrained for a segmentation task with RGB+XYZ data. Using these variants allows us to feed the model with many more points. Specifically, the samples are downsampled to $224 \times 224$, they are then reshaped into an unorganized point cloud (with the corresponding RGB values). For each point a $64$ dimensional features is returned. It is then reshaped back into an organized feature point cloud. As with the other methods, we pooled the features into $28 \times 28$ patches of $8 \times 8$ pixels. We use the code from the official github repository.

\subsection{SpinNet}
We use a model pretrained on 3DMatch \cite{3dmatch}. Prior to feeding the points into the network, we downsampled to $224 \times 224$. We then divided each sample into $28 \times 28$ patches of $8 \times 8$ pixels. These patches are then fed into the model which outputs a $32$ dimensional feature. We use the code from the official github repository.

\section{Preprocessing Implementation Details}
\label{app:preprocessing_impl}
\subsection{Plane Removal} By design, the objects in the dataset are centered within the image. We thus make a simplifying assumption that all the edges of the image lie on the same plane. To this end, we take a $10$ pixels wide strip around the image boundary from the organized point cloud. After removing all NaNs (i.e. noise), we use RANSAC \cite{ransac} to approximate the plane that best describes the boundary. The distance to this plane is calculated for each point in the point cloud, any point within $0.005$ distance is removed.
In practice, instead of removing the point, we zero the XYZ coordinates and RGB values for the point. This ensures the original resolution is kept. We use the Open3D \cite{open3d} "Segment Plane" implementations for the RANSAC step with $ransac\_n=50$ and $num\_iterations=1000$, for the actual plane removal we use the returned plane equation and manually zero the values.

\subsection{Clustering Based Outlier Removal} 
\begin{figure*}[t!]
    \centering
    \includegraphics[width=1\textwidth]{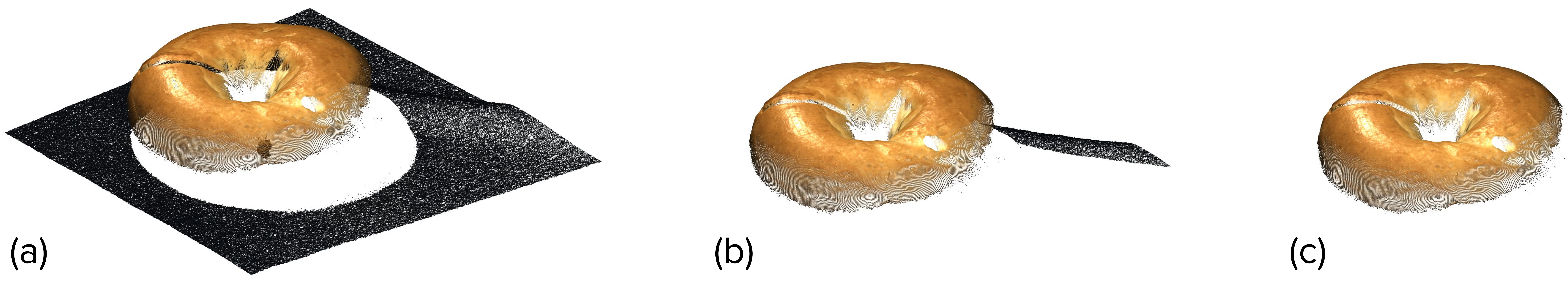}
    \caption{\textit{\textbf{3D-Aware Preprocessing:}} In (a), a nuisance artifact in the fabric. Although a RANSAC-based plane removal step approximates the best fitting plane, in some cases, artifacts are too far away from it to be removed (b). We use a connected components algorithm to discard the remaining artifacts (c)}
\label{fig:preprocessing}
\end{figure*}

Although the plane removal step can identify and remove the majority of the planes, in some cases, the planes are not planer, see Fig~\ref{fig:preprocessing}. Points in those areas may therefore be flagged as anomalies. By running DB-Scan \cite{dbscan} as a connected-components approach, each cluster is treated as a connected component. We keep the largest component and remove all points from other components (as before, we zero the XYZ coordinates and RGB values of the point).  We use the Open3D \cite{open3d} DB-Scan implementation with $\epsilon=0.006$ and $min\_points=30$.

\begin{table}[t]
\begin{center}
\caption{\textit{\textbf{Numeric Preprocessing results:}} average metrics across all classes are reported (Fig.~$7$-top in the main paper)}
\label{table:preprocessing_ablation}
\begin{tabular}{c@{\hskip5pt}c@{\hskip5pt}c@{\hskip5pt}c@{\hskip5pt}c@{\hskip5pt}c@{\hskip5pt}c@{\hskip5pt}c@{\hskip5pt}c@{\hskip5pt}} 
    \toprule
    & \multicolumn{2}{c}{RGB}  & \multicolumn{2}{c}{HoG} & \multicolumn{2}{c}{SIFT} & \multicolumn{2}{c}{FPFH}  \\
    \midrule
    & \textit{PRO} & \textit{I-ROC} & \textit{PRO} & \textit{I-ROC} & \textit{PRO} & \textit{I-ROC} & \textit{PRO} & \textit{I-ROC}\\
    \cmidrule(r){2-2}
    \cmidrule(r){3-3}
    \cmidrule(r){4-4}
    \cmidrule(r){5-5}
    \cmidrule(r){6-6}
    \cmidrule(r){7-7}
    \cmidrule(r){8-8}
    \cmidrule(r){9-9}
    
    Raw & \textbf{0.876} & \textbf{0.788} & 0.625 & 0.558 & 0.869 & 0.723 & \textbf{0.930} & 0.764  \\
    Pre & \textbf{0.876} & 0.770 & \textbf{0.771} & \textbf{0.559} & \textbf{0.910} & \textbf{0.727} & 0.924 & \textbf{0.782}  \\
  \bottomrule
    \end{tabular}
\end{center}
\end{table}

\end{document}